\PassOptionsToPackage{colorlinks,bookmarksopen,bookmarksnumbered,citecolor=red,urlcolor=red}{hyperref}
\documentclass[a4paper,fleqn]{cas-dc}

\usepackage[numbers]{natbib}
\usepackage{textcomp}
\usepackage{hyphenat}   
\usepackage{pifont} 
\usepackage{moreverb,url}
\usepackage{atbegshi}
\usepackage{multirow}
\usepackage[colorlinks,bookmarksopen,bookmarksnumbered,citecolor=red,urlcolor=red]{hyperref}
\usepackage{multirow}
\usepackage{tabularx}
\usepackage{soul}
\usepackage{xcolor}
\sethlcolor{lime}
\usepackage{booktabs}
\usepackage{enumitem}  
\usepackage{float}
\usepackage[utf8]{inputenc} 
\usepackage{newunicodechar}
\newunicodechar{≥}{\ensuremath{\geq}}
\newunicodechar{：}{:}
\usepackage[ruled,vlined]{algorithm2e}
\usepackage{amsmath}
\usepackage{atbegshi}
\def\tsc#1{\csdef{#1}{\textsc{\lowercase{#1}}\xspace}}
\tsc{WGM}
\tsc{QE}
\tsc{EP}
\tsc{PMS}
\tsc{BEC}
\tsc{DE}
\newcommand{\discardpages}[1]{%
  \begingroup
    \count0=#1
    \loop
      \AtBeginShipoutNext{\AtBeginShipoutDiscard}
      \advance\count0 -1
    \ifnum\count0>0 \repeat
  \endgroup
}
\begin{document}

\let\WriteBookmarks\relax
\def\floatpagepagefraction{1}
\def\textpagefraction{.001}
\shorttitle{Contrastive-Prototype Long-Tail CRS}
\shortauthors{Jinzhi Wang et~al.}

\title [mode = title]{LumiCRS: Asymmetric Contrastive Prototype Learning for Long-Tail Conversational Recommender Systems}                      
\tnotemark[1,2]
\tnotetext[1]{Code available at: \url{https://github.com/Jinzhi-Wang/LumiCRS-R1}
}
\tnotetext[2]{This work was supported by the National Natural Science Foundation of China under Grant No. 61872288.
}

\author[1]{Jinzhi Wang}[type=author,
                   auid=001,
                   bioid=1,
                   style=chinese,
                   orcid=0009-0008-8964-7164]
\ead{wangjz5515@stu.xjtu.edu.cn}

\credit{Conceptualization, Methodology, Formal analysis, Software, Data curation, Validation, Writing – original draft, Visualization, Writing – review \& editing}

\author[2]{Bin Li}[type=author,
                   auid=002,
                   bioid=2,
                   style=chinese,
                   orcid=0000-0002-6508-5071]
\ead{b.li2@siat.ac.cn}
\credit{Conceptualization, Supervision, Writing – review \& editing }
\author[1]{Qingke Peng}[type=author,
                   auid=003,
                   bioid=3,
                   style=chinese,
                   orcid=0000-0002-5448-8529]
\cormark[1]
\ead{qkpeng@mail.xjtu.edu.cn}
\credit{Conceptualization, Supervision, Methodology, Project administration, Writing – review \& editing, Funding acquisition}
\author[1]{Haozhou Li}[type=author,
                   auid=004,
                   bioid=4,
                   style=chinese]
\ead{lihaozhou1126@stu.xjtu.edu.cn}
\credit{Data curation, Software}
\author[1]{Zeyuan Zeng}[type=author,
                   auid=005,
                   bioid=5,
                   style=chinese]
\ead{zengzeyuan@stu.xjtu.edu.cn}
\credit{Data curation, Methodology, Writing – original draft}
\author[1]{Ruimeng Li}[type=author,
                   auid=006,
                   bioid=6,
                   style=chinese]
\ead{lrm1105@stu.xjtu.edu.cn}
\credit{Formal analysis, Validation}
\author[3]{Kaixuan Yang}[type=author,
                   auid=007,
                   bioid=7,
                   style=chinese]
\ead{yangkaixuan1@stu.xjtu.edu.cn}
\credit{Data curation, Visualization}
\author[4]{Jiangbo Zhang}[type=author,
                   auid=008,
                   bioid=8,
                   style=chinese]
\ead{4522253009@stu.xjtu.edu.cn}
\credit{Software, Visualization}
\author[1]{Biyi Zhou}[type=author,
                   auid=009,
                   bioid=9,
                   style=chinese]
\ead{zhoubiyi@stu.xjtu.edu.cn}
\credit{Investigation, Writing – review \& editing}
\author[5]{Yaoying Wang}[type=author,
                   auid=0010,
                   bioid=10,
                   style=chinese]
\ead{wangyaoying@stu.xjtu.edu.cn}
\credit{Writing – original draft, Validation}

\affiliation[1]{organization={Systems Engineering Institute, Xi’an Jiaotong University},
                addressline={No.28, Xianning West Road},
                city={Xi’an},
                postcode={710049},
                country={China}}
\affiliation[2]{organization={Shenzhen Institute of Advanced Technology, Chinese Academy of Sciences},
                addressline={1068 Xueyuan Avenue, Nanshan District},
                city={Shenzhen},
                postcode={518055},
                country={China}}
\affiliation[3]{organization={State Key Laboratory of Multiphase Flow in Power Engineering, School of Energy and Power Engineering, Xi’an Jiaotong University},
                addressline={No.28, Xianning West Road},
                city={Xi’an},
                postcode={710049},
                country={China}}
\affiliation[4]{organization={School of Electronic Science and Engineering, Xi'an Jiaotong University},
                addressline={No.28, Xianning West Road, Beilin District},
                city={Xi'an},
                postcode={710049},
                country={China}}

\affiliation[5]{organization={School of Software, Xi’an Jiaotong University},
                addressline={No.28, Xianning West Road, Beilin District},
                city={Xi’an},
                postcode={710049},
                country={China}}

\cortext[cor1]{Corresponding author}

\begin{abstract}
Conversational recommender systems (CRSs) often suffer from an extreme long-tail distribution of dialogue data, causing a strong bias toward head-frequency blockbusters that sacrifices diversity and exacerbates the cold-start problem. An empirical analysis of DCRS and statistics on the ReDial corpus show that only 10\,\% of head movies account for nearly half of all mentions, whereas about 70\,\% of tail movies receive merely 26\,\% of the attention. This imbalance gives rise to three critical challenges: head over-fitting, body representation drift, and tail sparsity. 
To address these issues, we propose LumiCRS, an end-to-end framework that mitigates long-tail imbalance through three mutually reinforcing layers: (i) an Adaptive Comprehensive Focal Loss~(ACFL) that dynamically adjusts class weights and focusing factors to curb head over-fitting and reduce popularity bias; (ii) Prototype Learning for Long-Tail Recommendation, which selects semantic, affective, and contextual prototypes to guide clustering and stabilize body and tail representations; and (iii) a GPT-4o-driven prototype-guided dialogue augmentation module that automatically generates diverse long-tail conversational snippets to alleviate tail sparsity and distribution shift. 
Together, these strategies enable LumiCRS to markedly improve recommendation accuracy, diversity, and fairness: on the ReDial and INSPIRED benchmarks, LumiCRS boosts Recall@10 and Tail-Recall@10 by 7--15\,\% over fifteen strong baselines, while human evaluations confirm superior fluency, informativeness, and long-tail relevance. These results demonstrate the effectiveness of multi-layer collaboration in building an efficient and fair long-tail conversational recommender.

\end{abstract}



\maketitle

\section{Introduction}
\begin{figure}
    \centering
    \includegraphics[width=1\linewidth]{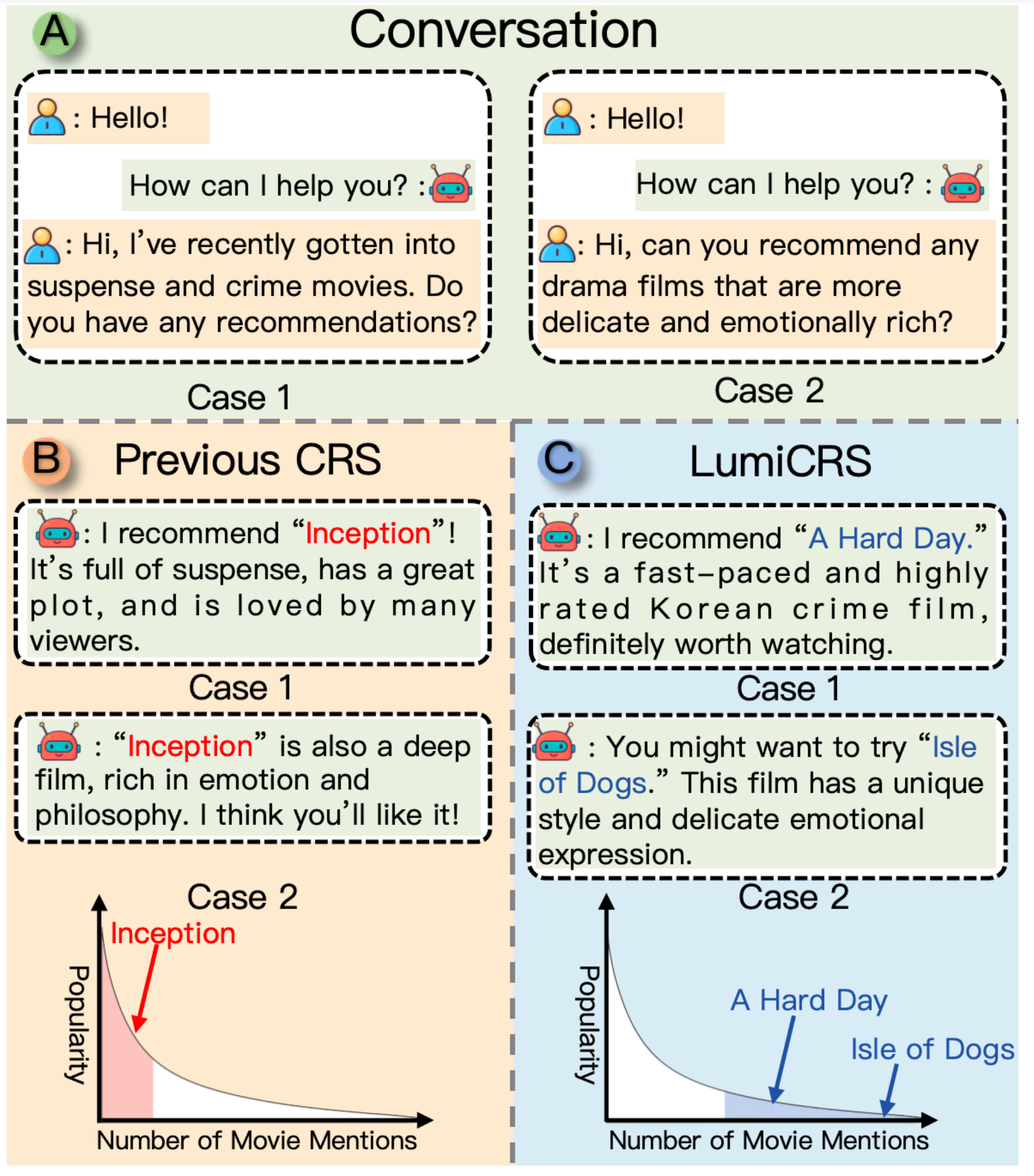}
    \caption{\textbf{Comparison of a conventional CRS with \textbf{LumiCRS} on two dialogue scenarios. } 
(A) Two users state distinct preferences: a suspense–crime thriller (Case 1) and a delicate, emotionally rich drama (Case 2).  
(B) The baseline CRS recommends the same head-frequency title \emph{Inception} in both cases, exposing a strong popularity bias.  
(C) \textbf{LumiCRS} instead surfaces niche yet relevant films—\emph{A Hard Day} and \emph{Isle of Dogs}—and their positions on the long-tail curve, illustrating effective long-tail modelling and more personalised, diverse recommendations.
}
    \label{introduction1}
\end{figure}

\begin{figure*}
    \centering
    \includegraphics[width=1\linewidth]{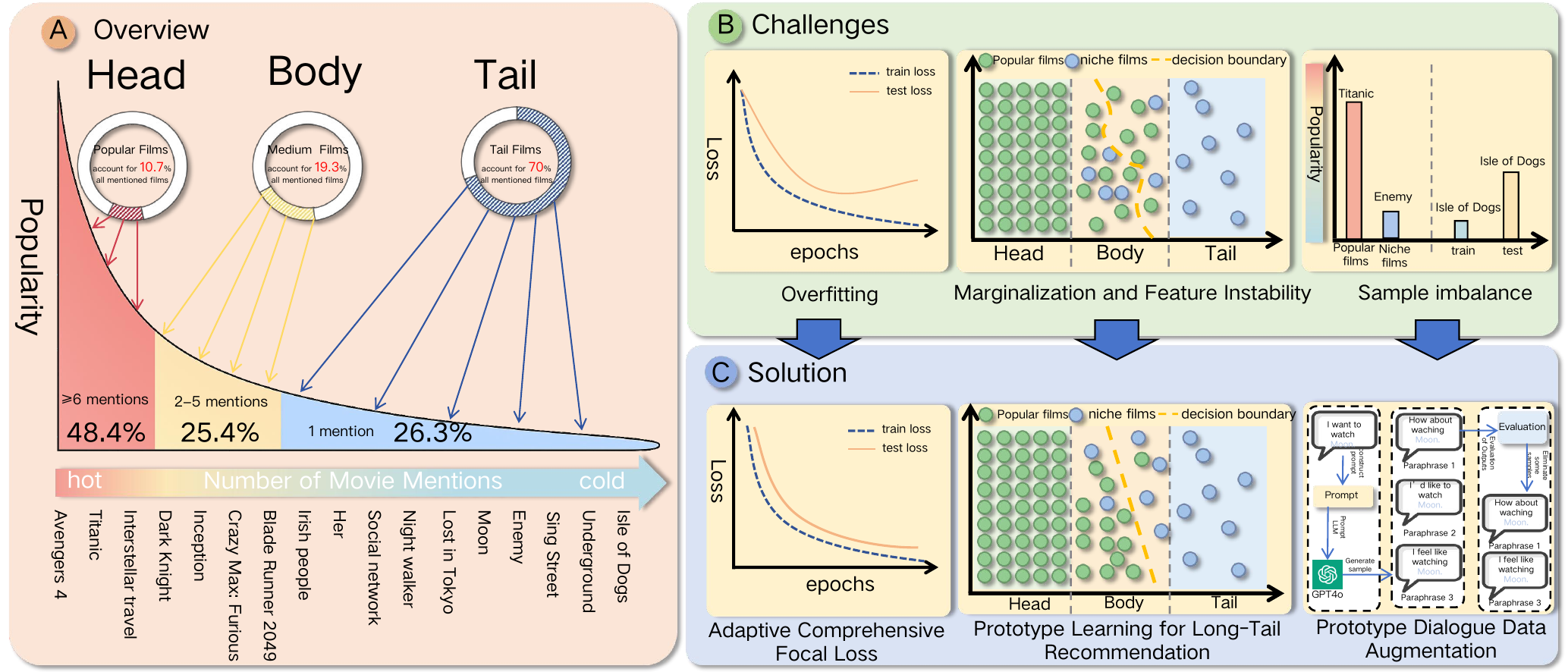}
    \caption{\textbf{Overview of the long-tail imbalance in \textsc{ReDial}, its key challenges, and our mitigation strategy.}
    (A) Head films ($\geq 6$ mentions) form only $10.7\%$ of titles but draw $48.4\%$ of mentions; Body ($2$--$5$) add $19.3\%/25.4\%$, while the Tail (single-mention, $\sim70\%$ of titles) receives just $26.3\%$. 
    (B) Head samples drive over-fitting, body samples experience feature drift, and tail samples are marginalised by the skewed distribution. 
    (C) Adaptive Comprehensive Focal Loss curbs head over-fitting; prototype learning stabilises body features for long-tail recommendation; GPT-4o dialogue augmentation enriches tail supervision.
    }
    \label{Introduction2}
\end{figure*}
Conversational Recommender Systems (CRSs) are multiturn natural language systems designed to capture users’ needs and preferences and deliver personalized, efficient recommendations \cite{jung2023towards}. A typical CRS consists of two main modules: an item recommendation component and a dialogue generation component. The item recommendation module identifies the user’s intent from the conversation and suggests appropriate items \cite{li2022user,ma2020cr}. The dialogue generation module produces natural, human-like responses to keep the interaction engaging and enhance user satisfaction  \cite{qian2023hutcrs,wu2022state}.

In recent years, CRS research has made significant strides. Early systems modeled user preferences solely from the conversation context \cite{bizer2009dbpedia,lu2021revcore}. However, since dialogue utterances are often brief and sparse in information, these early CRSs tended to fall back on recommending popular or generic items, hindering true personalization  \cite{chen2019towards}. To better capture user interests, later approaches enriched the context with external information sources such as knowledge graphs \cite{WU2023109873, speer2017conceptnet},hypergraphs \cite{LI2025113220,zheng2024hycorec}, product reviews \cite{zhou2022c2, ANWAR2025112579}, item metadata \cite{yang2021improving}, and related conversational data \cite{dao2024broadening}. The reasoning and language understanding capabilities of large pre-trained language models (e.g., Transformer-based models like RoBERTa) have also been leveraged to improve recommendation accuracy and contextual relevance \cite{zhou2020improving,wang2022towards}.

In parallel, growing privacy concerns have spurred the adoption of federated learning for CRSs, wherein conversational logs remain on-device and only lightweight model updates are exchanged. For example, FedRL \cite{di2024fedrl} and a diffusion-based recommender \cite{di2025federated, di2025personalized} achieved performance comparable to centralized models without transmitting raw user data. Researchers have also begun modeling users’ emotions and attitudes through multidimensional affective cues and explicit preference labels, further refining user representations and improving recommendation quality \cite{xi2024memocrs,xu2021adapting,zhao2023multi}. 
Overall, the field has progressed from simple context-based methods to more holistic approaches that incorporate rich external knowledge and advanced language models. Notable recent systems such as \mbox{C$^{2}$-CRS} \cite{zhou2022c2}, UniCRS \cite{wang2022towards}, and DCRS \cite{christakopoulou2016towards} employ these techniques to achieve significantly improved understanding of user queries and more fluent, relevant responses.

Despite these advances, significant challenges persist, especially in handling long-tail items \cite{FAREED2024112140, ZHANG2024111558}. Using the DCRS model  as a representative baseline, we observed recurring failures in scenarios where users requested niche content. Figure \ref{introduction1}A illustrates two sample dialogues: in one, a user requests a fast-paced crime thriller (case 1); in the other, a user requests a delicate emotional drama (case 2). As illustrated in Figure \ref{introduction1}B, a conventional CRS (DCRS) recommends the same blockbuster film, Inception, in both cases, demonstrating a strong bias toward head (high-popularity) movies. Even when users clearly articulate their specific preferences for niche films, the model tends to revert to mainstream titles, disregarding long-tail options. This behavior highlights the system's inadequate capability in modeling infrequent items and underscores the pronounced popularity bias inherent in such systems.

To investigate the root cause of this bias, we analyzed the distribution of movie mentions in the ReDial dialogue dataset. The data follows an extreme long-tail pattern (see Figure \ref{Introduction2}A): although head movies make up only about 10\% of the catalog, they account for roughly 48\% of all mentions. Body movies (moderately popular titles) constitute about 19\% of the catalog but receive around 25\% of mentions. In contrast, the tail movies — nearly 70\% of the entire movie pool — are mentioned only about 26\% of the time. In other words, a small number of blockbuster titles dominate nearly half of the conversations, while a vast number of lesser-known yet potentially relevant films are rarely brought up. This severe imbalance means that the CRS is trained far more on head items than on tail items, impairing its ability to learn users’ fine-grained preferences for niche content \cite{wang2025efficient}. As a result, recommendation diversity suffers and users are often steered toward the same popular options regardless of their unique tastes, undermining the personalization objective \cite{abdollahpouri2019managing}.

The inherent data imbalance of long-tail distributions
poses severe challenges to the two core tasks of CRSs—item
recommendation and dialogue generation (see Figure \ref{Introduction2}B) \cite{steck2018calibration}. To clarify the impact, we divide movies into three tiers according to their popularity \cite{abdollahpouri2017user}. First, the abundance of head-item examples can lead the model to overfit to popular movies and bias its recommendations toward these high-frequency items. Second, mid-frequency “body” items are under-represented in the training data, leading to unstable intermediate representations and inconsistent recommendation performance for these movies. Third, the extreme scarcity of tail-item interactions introduces severe train–test distribution mismatch and exacerbates the cold-start problem, making it difficult for the system to even recognize or suggest truly niche items. These issues compound throughout the recommendation pipeline: by the time the model generates a response, it relies excessively on content from head items, resulting in dialogues that lack diversity and personalization. In summary, current CRSs struggle to balance accuracy, diversity, and fairness due to the overwhelming popularity bias in their training data.

In this work, we propose LumiCRS, an end-to-end framework that addresses the long-tail bias through a coordinated three-tier strategy (see Figure \ref{Introduction2}C). At the loss-function level, we introduce an Adaptive Comprehensive Focal Loss (ACFL) that dynamically reweights the training process by down-weighting high-frequency (head) samples and adjusting focusing parameters. This prevents overfitting on popular items and mitigates their dominance in model learning. At the representation learning level, we develop a prototype-based module that selects representative prototypes based on semantic, emotional, and contextual similarity. These prototypes act as anchors to guide the model’s representation space, yielding clearer feature clusters and more stable representations for mid-frequency (body) items. At the data augmentation level, we design a prototype-guided dialogue augmentation mechanism powered by a GPT-4-based language model. This module automatically generates diverse new conversational snippets centered on tail movies (with an automated-manual validation pipeline to ensure quality), effectively enriching the training data for rare items. By integrating these three components, LumiCRS mitigates the long-tail imbalance at multiple levels.

For example, revisiting the earlier dialogues in Figure \ref{introduction1}C, LumiCRS can recommend a niche yet relevant title for each user’s request (A Hard Day for the crime thriller query and Isle of Dogs for the emotional drama query) instead of repeatedly suggesting a blockbuster like Inception. This outcome demonstrates that our approach provides more personalized and diverse recommendations by successfully modeling long-tail user preferences.

The main contributions of our work are summarized as follows:

\begin{enumerate}
    \item 
    We introduce Adaptive Comprehensive Focal Loss (ACFL), a dynamic loss function that rebalances training by down-weighting high-frequency samples. ACFL discourages the model from overfitting to head-heavy data and effectively mitigates popularity bias in recommendations.

    \item
   We propose a prototype-based representation learning method that clusters movies by semantic, affective, and contextual similarity. This approach improves the stability and discriminative quality of representations for mid-frequency “body” movies, leading to better recommendation performance on these often-neglected items.

    \item  
     We design a prototype-guided dialogue augmentation pipeline that leverages a large language model (GPT-4) to generate diverse conversational snippets for tail movies. By augmenting the training data for rare items, this method significantly increases tail-item coverage and enhances the system’s ability to handle sparsity, thereby alleviating cold-start issues in long-tail recommendation.
    \item 
    On the ReDial and \textsc{INSPIRED} datasets, LumiCRS sets new SOTA across all three tasks:  
Recall@10 improves by around 6.7\%, Tail-Recall@10 by around 10\%, Coverage@10 and ILD@10 by around 9\%, and Distinct-2 by around 10.5\%.  
In addition, with our newly proposed subjective evaluation metrics for conversational generation, LumiCRS also achieves significant improvements.
\end{enumerate}

\section{Related Work}
\subsection{CRS Architectures.}  
CRSs have evolved from early memory-based models to knowledge-enhanced approaches and, more recently, large language model (LLM)-based frameworks \cite{mb28}. Early CRSs relied solely on dialogue history to infer user preferences without external knowledge, often falling back on recommending popular items due to limited conversational signals \cite{harper2015movielens}. To overcome this limitation, knowledge-enhanced methods incorporated external resources such as knowledge graphs and topic models to enrich user modeling. Representative systems like KBRD \cite{chen2019towards}, KGSF \cite{zhou2020improving}, and ReDial \cite{li2018towards} leveraged structured knowledge to improve recommendation accuracy and response quality. However, these approaches heavily depend on the completeness and balance of external data, which may still bias the system toward head (popular) items. Recently, LLM-based methods have unified recommendation and dialogue generation by treating both as a joint language modeling task \cite{10778861}. Models such as UniCRS \cite{wang2022towards}, GUS   \cite{shi2025unifiedgenerativesearchrecommendation}, G-CRS  \cite{qiu2025graphretrievalaugmentedllmconversational}, and  MSCRS \cite{mb35} utilize large pre-trained models, sometimes augmented with knowledge or retrieved examples, to generate more coherent and informative responses. 

Although these methods demonstrate strong performance in dialogue fluency and flexibility, they generally inherit popularity bias from the training data and lack dedicated mechanisms or designs to address the long-tail issue, often favoring high-exposure mainstream items. Overall, despite incorporating external knowledge or advanced LLM techniques, existing conversational recommender systems still pay insufficient attention to niche content, with popular items continuing to dominate recommendation results.

\subsection{Long-Tail Recommendation Strategies.}  
Addressing long-tail item bias remains a key challenge in recommender systems, including conversational recommendation. In typical movie datasets, a small portion of head items (e.g., 10\%) often accounts for the majority of user interactions, while the vast number of tail items receive little to no exposure \cite{wang2025efficient}.  This imbalance reduces recommendation diversity and fairness, and it becomes even more severe in conversational scenarios where users rarely mention niche items. To mitigate long-tail bias, researchers have explored various strategies. One common approach involves adjusting loss functions to penalize popular items or favor rare ones, such as popularity-based re-weighting \cite{5680904}, inverse propensity scoring \cite{schnabel2016recommendationstreatmentsdebiasinglearning}, and causal inference methods \cite{Bonner_2018}. These techniques can improve long-tail coverage but often require careful hyperparameter tuning and may harm overall accuracy, especially as most were designed for static recommenders without considering the interactive dynamics of dialogues. Another strategy focuses on learning better representations for tail items through prototypical or contrastive learning, where methods like ProtoCF \cite{Sankar2021ProtoCF}, CDR \cite{10.1145/3627673.3679774}, CL4Rec \cite{Wei2022CL4Rec}, and ProtoGCD \cite{Liu2025ProtoGCD} have shown promise in offline long-tail retrieval tasks. 

However, these models generally ignore conversational contexts, assuming static user preferences, making it difficult to capture nuanced user intents expressed in dialogues, and they are rarely integrated into full conversational systems. Data augmentation offers a complementary solution by generating synthetic interactions for tail items, using techniques such as SimRec \cite{Gao2023SimRec}, ConvoAug \cite{Li2024ConvoAug}, and LLM-Aug \cite{Zhou2024LLMAug}. While these methods alleviate cold-start issues and expand training data, they also risk introducing noise and may cause models to overfit to artificial patterns. Moreover, most existing augmentation approaches focus on either recommendation or retrieval modules separately, lacking comprehensive end-to-end improvements for conversational recommendation.

In summary, while various long-tail mitigation techniques exist, they are typically designed as isolated solutions targeting specific components—such as re-ranking, re-weighting, or data augmentation—without addressing the complex interplay between conversational understanding and recommendation \cite{qin2021longtail}. Few approaches consider the multi-faceted nature of conversational recommendation, where dialogue dynamics, recommendation accuracy, and long-tail coverage must be balanced simultaneously. As a result, there remains a significant gap in developing holistic frameworks capable of jointly mitigating head-item bias, stabilizing mid-frequency item representations, and alleviating tail-item sparsity within a unified conversational recommender system. Our work aims to bridge this gap by integrating tailored solutions at the loss, representation, and data levels, collectively addressing long-tail bias while preserving dialogue quality.
\section{Preliminaries}
We denote the item set as \( I = \{i_1, i_2, \dots, i_N\} \), and the set of conversational contexts as \( S = \{s_1, s_2, \dots, s_M\} \). From the conversations in \( S \), we extract all mentioned entities to form the entity set \( E = \{e_1, e_2, \dots, e_K\} \), where \( I \subseteq E \). Here, \( N \) denotes the number of items, \( M \) the number of conversational contexts, and \( K \) the total number of entities.

Each conversational context \( s \in S \) consists of a sequence of utterances, denoted as:
\[
s = \{c_b\}_{b=1}^n.
\]
At the \( b \)-th turn, each utterance \( c_b \) is a sequence of words:
\[
c_b = \{w_j\}_{j=1}^m,
\]
where \( W \) denotes the vocabulary set.

As the conversation progresses, utterances are aggregated into a dialogue history. A Conversational Recommender System (CRS) leverages this history to infer user preferences and generate appropriate responses. At turn \( b \), the recommender module selects a set of candidate items from \( I \) based on the inferred preferences, while the dialogue module generates the next system utterance \( c_b \) based on the prior context.

\section{Methodology}
This section outlines our long-tail mitigation strategy across head, mid-tail, and tail items. Section 4.1 introduces an adaptive composite focused loss to curb head-item overfitting, whereas Sections 4.2 and 4.3 present a prototype-based representation framework and a GPT-4o-Based Dialogue Augmentation scheme to strengthen modeling of mid-tail and tail items, respectively. 
\subsection{Adaptive Comprehensive Focal Loss (ACFL)}
To combat the severe class imbalance and head-item overfitting induced by the long-tail phenomenon, we propose the Adaptive Comprehensive Focal Loss (ACFL). ACFL unifies class-frequency weighting, dynamic focusing factors, Top-K hard-example mining, and adaptive sampling within a single framework, thereby up-weighting tail samples, suppressing redundant high-frequency signals, and concentrating on hard instances during training. These mechanisms jointly mitigate popularity bias and substantially enhance both the accuracy and diversity of recommendations for long-tail items. This section explicates the design principles of ACFL, provides its complete mathematical formulation, and benchmarks it against recent state-of-the-art loss functions for long-tail recommendation and imbalanced classification.
\subsubsection{ACFL Formulation and Components}

We propose the Adaptive Comprehensive Focal Loss (ACFL), which substantially enhances the standard Focal Loss (FL) by addressing its inherent limitations. The original formulation of FL is as follows:
\begin{equation}
\text{FL}(p_t) = -\alpha_t (1 - p_t)^\gamma \log(p_t),
\end{equation}
where $p_t$ is the predicted probability of the true class, $\alpha_t$ denotes the class balance factor, and $\gamma$ is the focusing parameter controlling the down-weighting rate for easy examples. While effective, standard FL suffers from fixed hyperparameters, inadequate class imbalance consideration, and limited special handling of challenging samples.

To overcome these limitations, ACFL integrates four key mechanisms: adaptive class weighting, adaptive balancing and focusing factors, a Top-K selection strategy, and adaptive sampling. Specifically:

\textbf{Adaptive Class Weighting}: Class weights $w_c$ are dynamically adjusted according to class frequencies during training:
\begin{equation}
w_c = \frac{1}{N_c + \epsilon},
\end{equation}
where $N_c$ is the number of samples in class $c$, and $\epsilon$ is a smoothing factor.

\textbf{Adaptive Balance and Focus Factors}: Unlike standard FL with fixed parameters, ACFL dynamically adjusts the balance factor $\alpha_t$ and the focusing parameter $\gamma_t$ based on the difficulty of individual samples:
\begin{align}
\alpha_t &= \text{clip}(\alpha p_t + \epsilon, \alpha_{\min}, \alpha_{\max}), \\
\gamma_t &= \text{clip}(\gamma (1 - p_t), \gamma_{\min}, \gamma_{\max}).
\end{align}
Here, the \text{clip} function constrains the input to a fixed interval, i.e., $\text{clip}(x, a, b) = \min(\max(x, a), b)$, to prevent extremely small or large values from destabilizing training. These adjustments ensure that harder samples receive greater attention during model optimization.

\textbf{Top-K Hard Example Selection}: To further emphasize hard examples, ACFL employs a Top-K strategy that calculates the loss only for the most challenging samples. Given the predicted probability distribution, a threshold $\tau_k$ at the $k$-th percentile is used to define a binary mask $M_i$:
\begin{equation}
\tau_k = \text{quantile}(p_i, k), \quad M_i = \mathbf{1}\{p_i \geq \tau_k\}.
\end{equation}
Here, the \text{quantile} function returns the value below which a given percentage $k$ of the predicted probabilities $p_i$ fall. This ensures that only the top $(1-k)\%$ hardest examples (i.e., those with lower predicted confidence) are selected for loss computation.

\textbf{Adaptive Sampling Strategy}: An adaptive sampling method adjusts sampling weights based on class frequencies:
\begin{equation}
w_{\text{sample}, c} = \begin{cases}
1 + \frac{1}{N_c}, & N_c < \theta_{\min} \quad \text{(oversampling)}, \\
1 - \frac{N_c}{N_{\max}}, & N_c > \theta_{\max} \quad \text{(undersampling)}, \\
1, & \text{otherwise}.
\end{cases}
\end{equation}
\begin{algorithm}[htbp]
\caption{ACFL Full Procedure}
\label{alg:acfl_full}
\KwIn{
    Training samples $\{(x_j, y_j, c_j)\}_{j=1}^{N}$; \\
    Predicted probabilities $p_j = \hat{y}_j$; \\
    Class frequencies $\{N_c\}$; \\
    Popularity score $pop(x_j)$; \\
    Hyperparameters $\alpha, \gamma, \beta, k, \epsilon, \alpha_{\min}, \alpha_{\max}, \gamma_{\min}, \gamma_{\max}, \theta_{\min}, \theta_{\max}$
}
\KwOut{Total loss $\mathcal{L}_{\text{ACFL}}$}

\textbf{Step 1: Compute Class and Sample Weights}\;
\For{each class $c$}{
    Compute adaptive class weight: $w_c \gets \frac{1}{N_c + \epsilon}$\;
}
\For{$j \gets 1$ \KwTo $N$}{
    Assign sample weight: $w_{\text{sample},j} \gets$
    \begin{itemize}
        \item $1 + \frac{1}{N_{c_j}},$ \quad if $N_{c_j} < \theta_{\min}$ (oversampling)
        \item $1 - \frac{N_{c_j}}{N_{\max}},$ \quad if $N_{c_j} > \theta_{\max}$ (undersampling)
        \item $1,$ \quad otherwise
    \end{itemize}
}

\textbf{Step 2: Compute Adaptive Scaling Factors}\;
\For{$j \gets 1$ \KwTo $N$}{
    $\alpha_j \gets \text{clip}(\alpha \cdot p_j + \epsilon, \alpha_{\min}, \alpha_{\max})$\;
    $\gamma_j \gets \text{clip}(\gamma \cdot (1-p_j), \gamma_{\min}, \gamma_{\max})$\;
    Compute focal factor: $f_j \gets \alpha_j (1 - p_j)^{\gamma_j}$\;
    Compute popularity-based adjustment: $a_j \gets \exp(-\beta \cdot pop(x_j))$\;
    Compute weighted sample loss:\;
    \quad $l_j \gets -w_c \cdot w_{\text{sample},j} \cdot a_j \cdot f_j \cdot [y_j \log(p_j) + (1-y_j)\log(1-p_j)]$\;
}

\textbf{Step 3: Top-K Hard Example Selection}\;
Compute threshold $\tau_k \gets \text{quantile}(p_j, k)$\;
Define selection mask $M_j \gets \mathbf{1}\{p_j \geq \tau_k\}$\;

\textbf{Step 4: Aggregate Selected Losses}\;
Initialize total loss: $\mathcal{L}_{\text{ACFL}} \gets 0$\;

\For{$j \gets 1$ \KwTo $N$}{
    \If{$M_j = 1$}{
        $\mathcal{L}_{\text{ACFL}} \gets \mathcal{L}_{\text{ACFL}} + l_j$\;
    }
}
Normalize: $\mathcal{L}_{\text{ACFL}} \gets \frac{1}{\sum_j M_j} \cdot \mathcal{L}_{\text{ACFL}}$\;

\Return $\mathcal{L}_{\text{ACFL}}$
\end{algorithm}

\subsubsection{Comprehensiveness Verification of ACFL Design}
To rigorously evaluate the completeness and robustness of \textbf{ACFL} in long-tail recommendation, we benchmark it against six recent loss functions—Adaptive Logit Adjustment (ALA)~\cite{ALA2024}, Enhanced Adaptive Focal Loss (EAFL)~\cite{EAFL2024}, Asymmetric Loss with Padé Approximation (ALPA)~\cite{ALPA2025}, Cost-Sensitive Loss (CSL)~\cite{CSL2024}, Distributionally Robust Loss (DR-Loss)~\cite{DRLoss2024}, and Adaptive Hierarchical Loss (AHL)~\cite{AHL2025}. 
To highlight the unique advantages of ACFL in addressing extreme long-tail distributions, we compare its design against several recent loss functions across seven critical capabilities. These capabilities reflect essential aspects for long-tail modeling, including class weighting, hard-sample emphasis, adaptive parameter tuning, dynamic sampling, and others. As shown in Table~\ref{loss_table}, ACFL is the only method that simultaneously incorporates all seven properties, demonstrating its comprehensive suitability for highly imbalanced recommendation tasks.

\newcommand{\cmark}{\ding{51}}   
\newcommand{\xmark}{\ding{55}}   

\begin{table}[htbp]
\small         
\centering
\setlength\tabcolsep{4pt}  
\caption{\textbf{Comparison of Loss Functions on Key Capabilities for Long-Tail Recommendation.} 
We evaluate each method across seven essential properties for handling long-tail distributions: 
\textbf{CW} = Class Weighting, \textbf{IMB} = Imbalance Suitability, \textbf{ADA} = Adaptive Parameter Tuning, 
\textbf{HS} = Hard-Sample Emphasis, \textbf{NT} = No Manual Tuning, \textbf{DS} = Dynamic Sampling, 
\textbf{TK} = Top-K Hard Example Mining. }
\begin{tabular}{lccccccc}
\toprule
Method & CW & IMB & ADA & HS & NT & DS & TK\\
\midrule
ALA (2022)        & \cmark & \cmark & \cmark & \cmark & \xmark & \xmark & \xmark \\
EAFL (2024)       & \cmark & \cmark & \cmark & \cmark & \xmark & \xmark & \xmark \\
ALPA (2024)       & \cmark & \cmark & \cmark & \cmark & \xmark & \xmark & \xmark \\
CSL (2025)        & \cmark & \cmark & \cmark & \cmark & \cmark & \xmark & \xmark \\
DR‐Loss (2024)    & \cmark & \cmark & \cmark & \xmark & \xmark & \xmark & \xmark \\
AHL (2025)        & \cmark & \cmark & \cmark & \xmark & \xmark & \xmark & \xmark \\
\textbf{ACFL (Ours)} & \cmark & \cmark & \cmark & \cmark & \cmark & \cmark & \cmark \\
\bottomrule
\label{loss_table}
\end{tabular}
\end{table}

As Table~\ref{loss_table} indicates, ACFL is the only loss that fulfils all seven long-tail–oriented criteria, whereas each of the six baselines lacks at least two capabilities. Concretely, ACFL (i) suppresses head-class over-fitting and amplifies tail signals by coupling class-frequency weighting with a dynamic focusing factor, (ii) is the first to unify Top-$K$ hard-example mining and adaptive sampling within a single framework—thus providing simultaneous gradient- and data-level gains for tail learning—and (iii) employs a fully learnable, self-adjusting scheme that lets key hyper-parameters converge automatically as the distribution drifts. By contrast, ALA merely shifts logits and offers neither hard-sample selection nor adaptive resampling; EAFL improves tail sensitivity through a training-aware factor~$\gamma$ but relies on static distribution estimates and omits data-level sampling; ALPA supplies smooth asymmetric gradients suited to binary imbalance, yet is hard to extend to multi-class recommendation; CSL learns class costs via reinforcement learning, easing manual tuning yet incurring high computation and lacking gradient-based hard-example modelling; DR-Loss boosts worst-case robustness at the price of considerable overhead and no tail-focused mechanisms; and AHL equalises hierarchical gradients but requires extra redesign for flat, large-scale item spaces.

In summary, existing approaches either confine their adjustments to the loss layer—ignoring the need for tail-oriented reinforcement at both the data and gradient levels—or suffer from limitations in computational efficiency, cross-domain applicability, and automation. Consequently, they fall short of meeting the comprehensive demands of extreme long-tail scenarios. By contrast, ACFL’s complete coverage of all seven key dimensions makes it better suited than the above baselines for dialogue-based recommendation tasks that feature enormous item catalogs, severe data sparsity, and highly pronounced long-tail distributions.
\subsection{Prototype Learning for Long-Tail Recommendation}
\begin{figure*}
    \centering
    \includegraphics[width=1\linewidth]{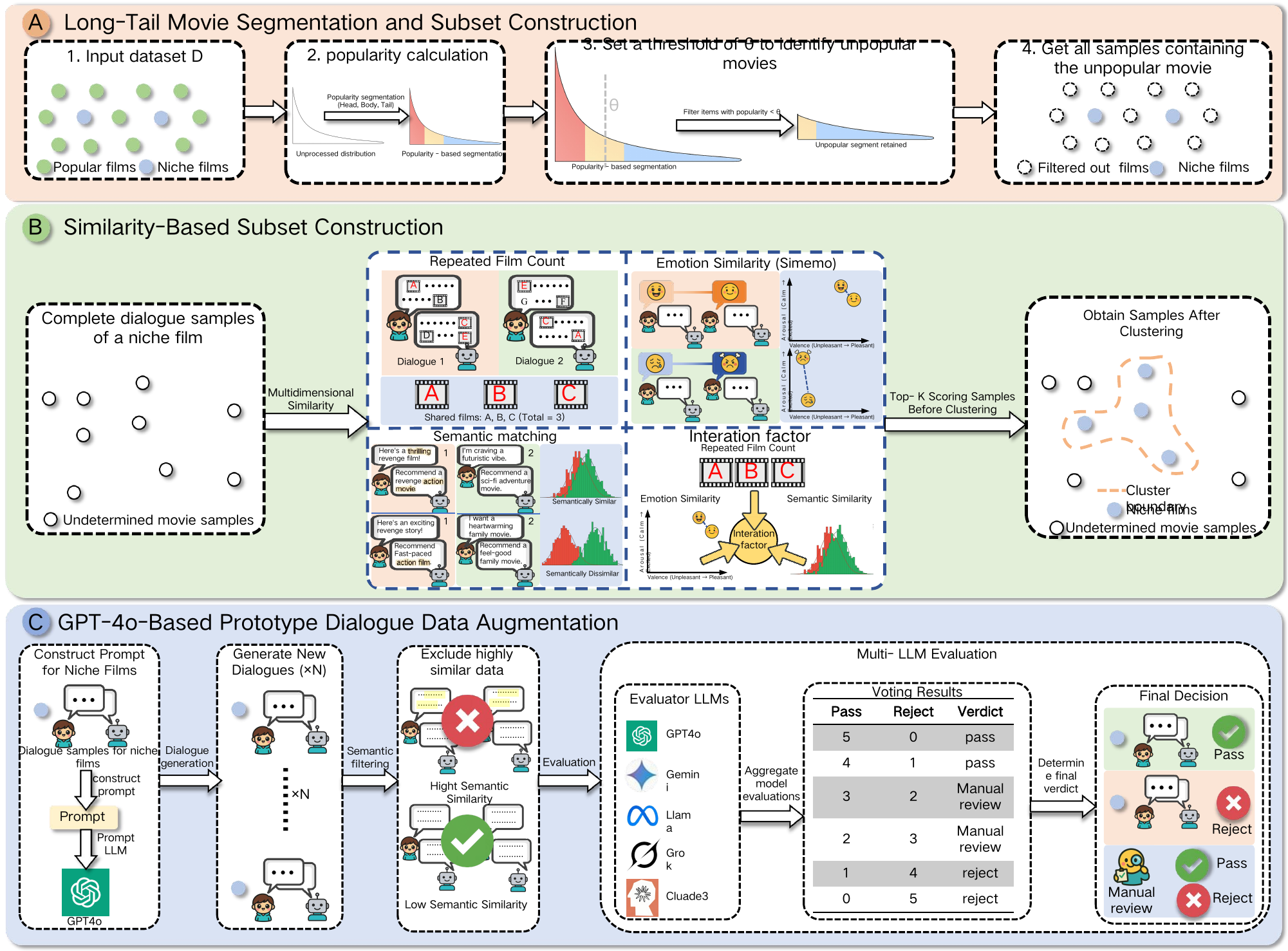}
    \caption{\textbf{Prototype construction and data-augmentation pipeline for mid- and long-tail movie dialogues.}
  (A) Segment the dataset by popularity and collect tail-related dialogues;
  (B) cluster high-quality prototypes using repetition counts plus affective and semantic similarity;
  (C) expand each prototype with GPT-4o-generated dialogues, filter near-duplicates, and validate the new data through multi-LLM voting with manual review.}
  \label{fig:pipeline}
    \label{Prototype Learning}
\end{figure*}
To address the head-body-tail imbalance in conversational movie recommendation, we propose a three-stage method that integrates prototype learning with data augmentation. This method focuses on the underrepresented body (mid-frequency) and tail (low-frequency) movies, while the head (high-frequency) movies are already well-covered by training data. 

By constructing representative prototypes for infrequent movies and enriching their context through clustering and augmentation, we obtain a high-quality training subset that improves coverage of mid- and long-tail items. The overall process (illustrated in Figure~\ref{Prototype Learning}) preserves semantic fidelity to the original data while greatly expanding the diversity of training samples for rare movies.

As a form of meta-learning, prototype learning is used to improve generalization in low-resource settings by learning from a small number of representative examples. Here, we leverage prototype learning to mitigate data sparsity for cold-start movies. Unlike naive clustering-based sampling that tends to favor frequent classes (i.e., majority movies), our prototype-based approach avoids majority-class dominance and enhances the visibility of low-frequency items.
In particular, we intentionally emphasize the body and tail segments of the item distribution, ensuring that both moderately popular and extremely rare movies receive adequate attention during training.
\subsubsection{Long-Tail Movie Segmentation and Subset Construction}

To address the challenges posed by the long-tail distribution in conversational recommendation, this section focuses on identifying mid-frequency and low-frequency (i.e., body and tail) movies from the training corpus and constructing structured data subsets for targeted modeling. The process, as illustrated in Figure~\ref{Prototype Learning}(a), involves two main steps: (i) popularity-based segmentation of movies into head, body, and tail groups; and (ii) data subset construction based on movie mentions in training samples.

\textbf{Popularity-Based Segmentation.}
We first compute the popularity of each movie as the number of training samples in which it is mentioned:
\[
\text{pop}(m) = |\{x \in D \mid x \text{ mentions movie } m\}|,
\]
where $D$ is the full training set. Based on the empirical distribution of popularity values, we define a threshold $\theta$ (e.g., the first quartile) to segment movies into three groups: movies with $\text{pop}(m) < \theta$ are classified as tail (low-frequency), those with moderate frequency (e.g., mentioned 2--5 times) form the body group, and the remaining frequent movies are assigned to the head group. This segmentation aligns with typical long-tail patterns and enables focused treatment of data-sparse regions.

\textbf{Subset Construction for Body and Tail Movies.}
After segmentation, we extract all training samples that mention each body or tail movie and organize them into hierarchical subsets as follows:

\begin{itemize}
    \item For each body movie $m \in M_{\text{body}}$, we collect its associated samples into a dedicated subset:
    \[
    D_m^{\text{body}} = \{ x \in D \mid x \text{ mentions movie } m \},
    \]
    \item These subsets are then aggregated to form the full body movie training set:
    \[
    D_{\text{body}} = \bigcup_{m \in M_{\text{body}}} D_m^{\text{body}},
    \]
    \item Likewise, for each tail movie $m \in M_{\text{tail}}$, we construct:
    \[
    D_m^{\text{tail}} = \{ x \in D \mid x \text{ mentions movie } m \},
    \]
    \item And merge them into the tail movie training set:
    \[
    D_{\text{tail}} = \bigcup_{m \in M_{\text{tail}}} D_m^{\text{tail}}.
    \]
\end{itemize}

Through this process, both $D_{\text{body}}$ and $D_{\text{tail}}$ contain only the training samples related to mid- and low-frequency movies, respectively. Each dataset is hierarchically organized into sub-collections per movie, providing a clear and structured foundation for subsequent modeling tasks such as prototype selection, similarity retrieval, and data augmentation.

\subsubsection{Similarity-Based Subset Construction}

Based on the previously constructed prototype set $C = \{ x^c_m \mid m \in M_{\text{tail}} \cup M_{\text{body}} \}$, we now retrieve additional training samples from the full corpus $D$ that are semantically or contextually similar to each prototype. This procedure yields an enriched training subset that enhances representation for long-tail movies, as illustrated in Figure~\ref{Prototype Learning}(b).

\subsubsection*{Composite Similarity Function}

We define a composite scoring function $\mathrm{FinalScore}(x^c_m, x)$ to evaluate the similarity between a prototype $x^c_m$ and a candidate sample $x \in D$. This function integrates four key components:

\begin{itemize}
    \item \textbf{Semantic Similarity} ($\mathrm{Sim}_{\text{sem}}$):  
    Let $v_{x}$ denote the text-based feature vector (e.g., embedding) of sample $x$. Semantic similarity is measured by cosine similarity:
    \[
    \mathrm{Sim}_{\text{sem}}(x^c_m, x) = \frac{v_{x^c_m} \cdot v_{x}}{\|v_{x^c_m}\| \cdot \|v_{x}\|}
    \]

    \item \textbf{Emotional Similarity} ($\mathrm{Sim}_{\text{emo}}$):  
    Let $e_{x}$ denote the emotional feature vector of sample $x$. We compute emotional similarity using inverse L1 distance:
    \[
    \mathrm{Sim}_{\text{emo}}(x^c_m, x) = \frac{1}{1 + \|e_{x^c_m} - e_x\|_1}
    \]

    \item \textbf{Residual Movie Overlap} ($\mathrm{Sim}_{\text{mov}}$):  
    Let $M(x)$ be the set of movies mentioned in sample $x$. Residual movie overlap is measured by the cardinality of shared movies:
    \[
    \mathrm{Sim}_{\text{mov}}(x^c_m, x) = |M(x^c_m) \cap M(x)|
    \]

    \item \textbf{Interaction Factor} ($R$):  
    This factor combines the above metrics non-linearly:
    \[
    \begin{aligned}
    R(x^c_m, x) &= \mathrm{Sim}_{\text{sem}}(x^c_m, x) \cdot \mathrm{Sim}_{\text{emo}}(x^c_m, x) \\
    &\quad \cdot \frac{1}{1 + \exp\bigl(-\alpha \cdot \mathrm{Sim}_{\text{mov}}(x^c_m, x)\bigr)}
    \end{aligned}
    \]
    where $\alpha$ is a hyperparameter controlling the weight of residual movie matching.
\end{itemize}

\noindent The final similarity score is a weighted sum of all components:
\[
\begin{aligned}
\mathrm{FinalScore}(x^c_m, x) 
&= w_{\text{sem}} \cdot \mathrm{Sim}_{\text{sem}}(x^c_m, x) \\
&\quad + w_{\text{emo}} \cdot \mathrm{Sim}_{\text{emo}}(x^c_m, x) \\
&\quad + w_{\text{mov}} \cdot \mathrm{Sim}_{\text{mov}}(x^c_m, x) \\
&\quad + w_{\text{int}} \cdot R(x^c_m, x)
\end{aligned}
\]
where $w_{\text{sem}}, w_{\text{emo}}, w_{\text{mov}}, w_{\text{int}}$ are predefined or learnable weights normalized to sum to 1.

\textbf{Enriched Subset Construction.}  
For each prototype $x^c_m$, we rank all samples $x \in D$ by $\mathrm{FinalScore}(x^c_m, x)$ and select the top-$K$ neighbors as a support set:
\[
S_m = \text{Top-}K\text{Neighbors}(x^c_m),
\]
and aggregate them to form the overall training subset for long-tail enhancement:
\[
S_1 = \bigcup_{m \in M_{\text{tail}} \cup M_{\text{body}}} S_m.
\]

The resulting subset $S_1$ consists of both prototype samples and their most relevant contextual neighbors, including ``bridge'' samples from head movies when they are highly similar in semantics or emotion. This construction mitigates head-dominance and provides a rich, balanced context for tail and body movies in downstream training.
\subsection{GPT-4o-Based Prototype Dialogue Data Augmentation}

In the GPT-4o-Based Prototype Dialogue Data Augmentation module, we tackle the lingering data-sparsity problem of mid- and low-frequency movies by executing a four-stage pipeline (Figure\ref{fig:pipeline}.C) that complies with the notation and structure of the preceding methodology. First, prototype dialogues are embedded into task-specific prompt templates; next, GPT-4o performs large-model generation, producing a moderate batch for body movies and an intensified batch for tail movies. All outputs then pass through semantic filtering with multi-model evaluation, where SimCSE similarity pruning is followed by majority voting from five diverse LLMs. Finally, human review and data integration audit edge cases and merge the accepted dialogues—tagged by their body or tail origin—into the training corpus, delivering rich yet quality-controlled augmentation.

\textbf{1.~Prompt Template Construction.}  
For each body or tail movie, we concatenate its prototype sample with semantically similar dialogues to build a contextual prompt, which instructs GPT-4o to generate recommendation dialogues that explicitly mention the target movie.

\textbf{2.~Large-Model Generation.}  
GPT-4o generates multi-turn, diverse candidate dialogues under a high sampling temperature (e.g., 0.8). Output volume is frequency-aware:  
\begin{itemize}
  \item Tail movies: \textbf{ 8--10} dialogues per movie.  
  \item Body movies: \textbf{ 4--5} dialogues per movie.  
\end{itemize}

\textbf{3.~Semantic Similarity Filtering.}  
We compute Sentence-BERT cosine similarity between each generated dialogue and its prototype. Candidates with similarity~$>0.85$ are discarded, retaining only those that preserve semantic fidelity while adding linguistic variety.

\textbf{4.~Multi-Model Automatic Evaluation \& Human Review.}  
Each surviving candidate is independently scored by five leading LLMs (GPT-4o, Gemini, LLaMA, Grok, Claude~3) on semantic consistency, fluency, and recommendation plausibility:  
\begin{itemize}
  \item \textbf{4–5} models pass $\Rightarrow$ sample automatically accepted.  
  \item \textbf{2–3} models pass $\Rightarrow$ forwarded to human annotators.  
  \item \textbf{0–1} models pass $\Rightarrow$ sample discarded.  
\end{itemize}
Human reviewers apply consistent criteria to borderline cases, ensuring only high-quality dialogues remain.

\textbf{5.~Data Integration.}  
All accepted dialogues are merged with the original corpus to create an expanded training set: tail movies receive substantial augmentation, body movies moderate expansion, and head movies remain unchanged. This stratified strategy enhances long-tail coverage while preserving language quality and recommendation relevance, providing a robust foundation for subsequent training.
\section{Experimental Setup}
\subsection{Dataset}
To evaluate the effectiveness and generalizability of our method in the recommendation of conversational movies –particularly in long-tail and data-sparse settings –we adopt two benchmark datasets: \textbf{ReDial} and \textbf{INSPIRED}. These datasets differ in dialogue style, user preference expression, and item frequency distribution. Statistics are summarized in Table~\ref{tab:longtail_stats}.

\begin{itemize}
    \item \textbf{ReDial}~\cite{li2018towards} is a widely used dataset collected via Amazon Mechanical Turk (AMT). It includes user-like/dislike feedback and supports fundamental preference modeling.
    \item \textbf{INSPIRED}~\cite{hayati2020inspired} is designed to explore persona-grounded recommendations, with shorter dialogues and more consistent structure.
\end{itemize}
\begin{table}[ht]
\centering
\small
\setlength{\tabcolsep}{5pt}
\caption{\textbf{Long-tail distribution of movies in each dataset.} Each cell shows percentage and count of movies in Head, Body, and Tail groups.}
\label{tab:longtail_stats}
\begin{tabular}{lccc>{\centering\arraybackslash}p{1cm}>{\centering\arraybackslash}p{1cm}>{\centering\arraybackslash}p{1cm}}
\toprule
\textbf{Dataset} & \textbf{Dialogs} & \textbf{Movies} & \textbf{Head} & \textbf{Body} & \textbf{Tail} \\
\midrule
ReDial   & 31,089 & 5,896 & 10.7\%   & 19.3\% & 70.0\% \\
INSPIRED & 1,997  & 1,058 & 11.3\%   & 28.4\%  & 60.3\%   \\
\bottomrule
\end{tabular}
\end{table}

\subsection{Baseline}
Following the survey and benchmarking protocol introduced by Dao et al.~\cite{mb35}, we compare our model against 15 representative CRS baselines, covering four technical paradigms: statistical ranking, pre-trained language models (PLMs), knowledge graph (KG)–enhanced architectures, and retrieval-based or multi-strategy frameworks. To ensure fair comparison, all implementations and hyperparameters are reproduced from publicly available code.

\begin{itemize}[leftmargin=*]
    \item \underline{\textbf{Popularity}}: A simple baseline that ranks items based on how frequently they appear in the dataset. It provides a non-personalized reference for evaluation.
    
    \item \underline{\textbf{TextCNN}} \cite{mb39}:  It applies convolutional neural networks to capture local contextual features in dialogue and ranks candidate movies accordingly.
    
    \item \underline{\textbf{BERT}} \cite{mb40}:  A widely adopted pre-trained encoder designed for text classification tasks. We fine-tune BERT to predict relevant items from user utterances.
    
    \item \underline{\textbf{GPT-2}} \cite{mb41}: It serves as a large-scale autoregressive language-model baseline for both context modelling and text generation.
    
    \item \underline{\textbf{DialoGPT}} \cite{mb42}: It fine-tunes GPT-2 on extensive conversational data to generate fluent and contextually coherent responses.
    
    \item \underline{\textbf{BART}} \cite{mb43}: A denoising autoencoder architecture that supports sequence-to-sequence generation, used here for both recommendation reasoning and natural response construction.
    
    \item \underline{\textbf{ReDial}} \cite{mb28}: It combines an auto-encoder recommender with an HRED generator, forming the earliest English CRS baseline.
    
    \item \underline{\textbf{KBRD}} \cite{mb24}: The framework introduces a structured knowledge graph to enhance the entity-level understanding of both recommendation and generation.
    
    \item \underline{\textbf{KGSF}} \cite{zhou2020improving}: It integrates word-level and entity-level knowledge graphs and aligns them via mutual-information maximisation to enrich representations.
    
    \item \underline{\textbf{UniCRS}} \cite{wang2022towards}: It unifies recommendation and dialogue subtasks within a prompt-learning paradigm driven by large pre-trained language models.
    
    \item \underline{\textbf{TREA}} \cite{mb29}:  A reasoning-based system that leverages a multi-layer inferable entity tree to guide conversational flow and recommendation logic.
    
    \item \underline{\textbf{VRICR}} \cite{mb31}: A variational inference-based framework that completes sparse knowledge graphs in real time, adapting to dynamic dialogue contexts.
    
    \item \underline{\textbf{DCRS}} \cite{mb35}: This model retrieves demonstration dialogues highly similar to the current context, guiding both recommendation and response generation in an in-context learning style.
    
    \item \underline{\textbf{DisenCRS}} \cite{disencrs}: An information-theoretic model that disentangles user preferences via contrastive learning, targeting improved long-tail recall and intent clarity.
    
    \item \underline{\textbf{MSCRS}} \cite{mscrs}: A two-stage approach that first retrieves and then refines candidate responses to balance recommendation precision with conversational diversity.
\end{itemize}

\subsection{Evaluation Metrics}
In conversational recommender systems (CRSs), the workflow is divided into two core tasks—item recommendation and response generation—which correspond to the research questions RQ1 and RQ2 in this paper. To provide a comprehensive assessment of model performance, we establish a systematic set of evaluation metrics for each task, enabling us to quantify overall accuracy while capturing behavior in long-tail scenarios and from a user-experience perspective.

\begin{table}[htbp!]
\caption{\textbf{Detailed scoring criteria for metrics.}Each metric is rated on a five-point Likert scale (1 = lowest, 5 = highest); the second column lists the qualitative criteria used to assign the scores.}
\label{Metrics}
\centering
\scriptsize
\begin{tabular}{p{1.5cm}p{6cm}}
\toprule
\multicolumn{1}{c}{\textbf{Metrics}} & \multicolumn{1}{c}{\textbf{Scoring details}}\\
\midrule
Logical Persuasiveness & 1 point: Argument·s are incoherent and unconvincing.

2 points: 	Arguments are discernible but lack rigor.

3 points: Arguments are generally coherent with occasional logical gaps.

4 points: Arguments are clear, well‑reasoned, and persuasive.

5 points: Arguments are exceptionally rigorous and compelling, fully convincing the user.\\
\hline
\vspace{5pt} 
Informativeness &1 point: Provides no useful information.

2 points: Provides minimal, shallow information.

3 points: Provides a moderate amount of factual and contextual information.

4 points: Provides rich, detailed, and practical information.

5 points: Provides comprehensive, in‑depth, and highly practical information.\\
\hline
\vspace{5pt} 
Lifelikeness & 1 point:Responses are mechanical and devoid of human touch.

2 points:Slightly natural but still noticeably stiff.

3 points: Responses are fluent and somewhat engaging

4 points: Responses are vivid, engaging, and immersive.

5 points:Responses are highly lifelike, emotionally resonant, and akin to real human conversation.\\
\hline
\vspace{5pt} 
Long‑Tail Relevance & 1 point: Never mentions or recommends any niche movies.

2 points: Occasionally mentions niche movies, but in a superficial or irrelevant manner

3 points: Consistently recommends some niche movies, meeting basic long‑tail discovery needs.

4 points: Recommends several highly relevant niche/low‑frequency movies with good coverage.

5 points: Demonstrates mastery of long‑tail content by precisely recommending numerous high‑quality niche films. \\\hline
\vspace{5pt} 
Recommendation Diversity & 1 point: Recommendations consist of the same popular title with no variety. 

2 points: Includes a few different titles but remains dominated by popular items.

3 points: Covers multiple genres or styles, partially overcoming popular‑item repetition bias.

4 points:Shows clear diversity in genre and style, effectively avoiding repetition of popular titles.

5 points: Delivers a highly diverse recommendation list covering a wide range of categories and styles, maximizing user surprise. \\
\bottomrule
\end{tabular}
\end{table}

For item recommendation (RQ1), evaluation focuses on two dimensions: accuracy and long-tail coverage.
Accuracy is measured with Recall@k (k = 1, 10, 50), NDCG@k (k = 10, 50), and MRR@k (k = 10, 50). Recall@k verifies whether the ground-truth movie appears within the top-k results; NDCG@k further discounts by rank, rewarding relevant items that are ranked higher; MRR@k considers only the position of the first relevant item, reflecting the model’s ability to hit the target as early as possible.
Long-tail and diversity are captured by Tail-Recall@k, Coverage@k, ILD@k, and PWP. Tail-Recall@k records the proportion of low-frequency movies correctly retrieved in the top-k list; Coverage@k measures how much of the entire catalogue is covered by the top-k recommendations; ILD@k (Intra-List Diversity) quantifies topical diversity within the list; PWP (Popularity-Weighted Precision) down-weights popular items to diagnose popularity bias. Together, these seven metrics provide a holistic view of both overall accuracy and fairness/diversity in the long-tail.

For response generation (RQ2), evaluation is conducted on two levels: objective automatic metrics and subjective human metrics.
On the objective side, we adopt BLEU-2 and BLEU-3 to gauge n-gram precision and thus lexical fluency; ROUGE-L (based on the longest common subsequence) to measure content coverage; and DIST-2, DIST-3, and DIST-4 (Distinct n-gram Ratios) to quantify lexical diversity.
On the subjective side, we employ five 0–5 scales—logical persuasiveness (Log Pers), informativeness (Info), liveliness (Life), long-tail relevance (Tail Rel), and diversity (Div)—with detailed rubrics provided in Table\ref{Metrics}. Log Pers assesses the coherence and rigor of the system’s reasoning; Info measures the richness of facts, context, and recommendation rationale; Life evaluates the naturalness and immersive quality of responses; Tail Rel reflects the ability to surface niche, low-frequency movies; and Div gauges genre and stylistic breadth in the recommendation list.
The evaluation procedure is as follows: we randomly sample 1 000 system responses and score them automatically with GPT-4 Turbo. From these, 200 responses are further evaluated independently by three expert annotators. Correlations between automatic and human scores are then computed to ensure the reliability of the subjective assessment.
\subsection{ Implementation Details}
We implement LumiCRS with PyTorch. We fine-tuned DialoGPT on our dataset for 8 hours using 3 * 3090 GPUs with a learning rate of 5e-05, 20 training epochs, and a batch size of 12. For the Llama2chat, we conducted 60 hours of Qlora parameter fine-tuning on a dataset using 3 * 3090 GPUs with a learning rate of 1e-4, 20 training epochs, and a batch size of 10. We also specified Lora layer parameters with a rank of 64, an alpha value of 16, and a dropout rate of 0.05.
\section{ Experiments}
\begin{table*}[htbp]
\caption{\textbf{Recommendation performance comparison on ReDial and INSPIRED datasets.} Bold “*” = best with $p<0.05$;   \underline{value} = strongest baseline;   Gain = relative improvement over the strongest baseline.}
\label{tab:rec}
\centering
\resizebox{\textwidth}{!}{
\begin{tabular}{c|ccc|cc|cc|ccc|cc|cc}
\toprule
 & \multicolumn{7}{c|}{\textbf{ReDial}} & \multicolumn{7}{c}{\textbf{INSPIRED}} \\
\cmidrule(lr){2-8} \cmidrule(lr){9-15}
\multirow{1}{*}{\textbf{Model}} & \multicolumn{3}{c|}{\textbf{Recall}} & \multicolumn{2}{c|}{\textbf{NDCG}} & \multicolumn{2}{c|}{\textbf{MRR}} & \multicolumn{3}{c|}{\textbf{Recall}} & \multicolumn{2}{c|}{\textbf{NDCG}} & \multicolumn{2}{c}{\textbf{MRR}} \\
\cmidrule(lr){2-4} \cmidrule(lr){5-6} \cmidrule(lr){7-8} \cmidrule(lr){9-11} \cmidrule(lr){12-13} \cmidrule(lr){14-15}
 & \textbf{@1} & \textbf{@10} & \textbf{@50} & \textbf{@10} & \textbf{@50} & \textbf{@10} & \textbf{@50} & \textbf{@1} & \textbf{@10} & \textbf{@50} & \textbf{@10} & \textbf{@50} & \textbf{@10} & \textbf{@50} \\
\midrule
Popularity     & 0.011 & 0.053 & 0.183 & 0.029 & 0.057 & 0.021 & 0.027 & 0.031 & 0.155 & 0.322 & 0.085 & 0.122 & 0.064 & 0.071 \\
TextCNN   & 0.010 & 0.066 & 0.187 & 0.033 & 0.059 & 0.023 & 0.028 & 0.025 & 0.119 & 0.245 & 0.066 & 0.094 & 0.050 & 0.056 \\
BERT       & 0.027 & 0.142 & 0.307 & 0.075 & 0.112 & 0.055 & 0.063 & 0.049 & 0.189 & 0.322 & 0.112 & 0.141 & 0.088 & 0.095 \\
\midrule
ReDial    & 0.010 & 0.065 & 0.182 & 0.034 & 0.059 & 0.024 & 0.029 & 0.009 & 0.048 & 0.213 & 0.023 & 0.059 & 0.015 & 0.023 \\
KBRD     & 0.033 & 0.150 & 0.311 & 0.083 & 0.118 & 0.062 & 0.070 & 0.042 & 0.135 & 0.236 & 0.088 & 0.109 & 0.073 & 0.077 \\
KGSF      & 0.035 & 0.175 & 0.367 & 0.094 & 0.137 & 0.070 & 0.079 & 0.051 & 0.132 & 0.239 & 0.092 & 0.114 & 0.079 & 0.083 \\
TREA    & 0.045 & 0.204 & 0.403 & 0.114 & 0.158 & 0.087 & 0.096 & 0.047 & 0.146 & 0.347 & 0.095 & 0.132 & 0.076 & 0.087 \\
COLA   & 0.048 & 0.221 & 0.426 & - & - & 0.086 & 0.096 & - & - & - & - & - & - \\
VRICR    & 0.054 & 0.244 & 0.406 & 0.138 & 0.174 & 0.106 & 0.114 & 0.043 & 0.141 & 0.336 & 0.091 & 0.134 & 0.075 & 0.085 \\
UNICRS   & 0.065 & 0.241 & 0.423 & 0.143 & 0.183 & 0.113 & 0.125 & 0.085 & 0.230 & 0.398 & 0.149 & 0.187 & 0.125 & 0.133 \\
DCRS & 0.076& 0.253 & 0.439 & 0.154 & 0.195 & 0.123 & 0.132 & 0.093 & 0.226 & 0.414 & 0.153 & 0.192 & 0.130 & 0.137 \\
MSCRS  & 0.081 & 0.264 & 0.451 & 0.161 & 0.201 & 0.128 & 0.136 & \underline{0.096} & \underline{0.257} & \underline{0.425} & \underline{0.168} & \underline{0.202} & \underline{0.140} & \underline{0.148} \\
DisenCRS  & \underline{0.081} & \underline{0.268} & \underline{0.451} & \underline{0.162} & \underline{0.210} & \underline{0.130} & \underline{0.138} & 0.094 & 0.252 & 0.423 & 0.165 & 0.200 & 0.139 & 0.146 \\
\rowcolor{gray!10}
\textbf{LumiCRS}  & \textbf{0.088*} & \textbf{0.286*} & \textbf{0.495*} & \textbf{0.173*} & \textbf{0.220*} & \textbf{0.139*} & \textbf{0.149*} & \textbf{0.104*} & \textbf{0.274*} & \textbf{0.466*} & \textbf{0.179*} & \textbf{0.212*} & \textbf{0.150*} & \textbf{0.160*} \\
\midrule
\textit{Gain (\%)} & 7.95\% & 6.72\% & 9.76\% & 6.79\% & 4.76\% & 6.92\% & 7.97\% & 8.33\% & 6.61\% & 9.65\% & 6.55\% & 4.95\% & 7.14\% & 8.11\% \\
\bottomrule
\end{tabular}
}
\end{table*}
To comprehensively demonstrate the effectiveness of our proposed LumiCRS, we design and conduct extensive experiments to answer the following key questions:

\textbf{RQ1:} How does LumiCRS compare with existing baselines in terms of recommendation accuracy and long-tail effectiveness?

\textbf{RQ2:} How does LumiCRS compare with existing baselines in terms of dialogue-generation quality and diversity?

\textbf{RQ3:} What is the impact of each core module in LumiCRS?

\textbf{RQ4:} Can the Adaptive Comprehensive Focal Loss (ACFL) effectively alleviate overfitting and improve generalization performance during training?

\textbf{RQ5:} Can prototype learning improve the representation quality and robustness for mid-frequency movies?

\textbf{RQ6:} Can prototype-driven augmentation improve long-tail and cold-start recommendation performance?

\textbf{RQ7:} How sensitive is LumiCRS to its key hyperparameters in ACFL, prototype learning, and data augmentation, and how do they affect overall and long-tail performance?

\subsection{Evaluation on Recommendation Task (RQ1)}

We evaluate the recommendation performance of LumiCRS on both the ReDial and INSPIRED datasets, focusing on two major dimensions: overall ranking accuracy and long-tail recommendation ability (complete results are presented in Table~\ref{tab:rec} and Table~\ref{tab:longtail}). Regarding overall ranking accuracy, LumiCRS significantly outperforms the strongest baselines across all ranking metrics (Recall, NDCG, MRR at k = 1, 10, 50) on both datasets. For instance, on ReDial, LumiCRS improves Recall@1/10/50 by 7.95\%, 6.72\%, and 9.76\% over DisenCRS, respectively. Similarly, all four NDCG and MRR indicators achieve gains of 5\%–8\%. On the sparser INSPIRED dataset, LumiCRS still attains improvements of 8.33\%, 6.61\%, and 9.65\% in Recall@1/10/50, demonstrating its ability to accurately capture both immediate and overall user preferences across different ranking lengths.

Compared to the overall metrics, the long-tail recommendation capability of LumiCRS further highlights its advantages. In ReDial, LumiCRS raises Tail-Recall @1/10/50 0.024/0.067/0.114, corresponding to relative gains of 14.29\%, 9.84\%, and 11.76\%, respectively. Furthermore, Coverage@10 /50 and ILD@10/50 show improvements close to 10\%, indicating that LumiCRS not only hits more low-frequency movies but also significantly expands the exposure range of tail items and diversifies the recommendation list. On the more sparse INSPIRED dataset, the long-tail benefits of LumiCRS become even more pronounced: Tail-Recall@1/10/50 improves by 15.79\%, 10.71\%, and 12.90\% compared to the baseline, while Coverage and ILD metrics consistently maintain a 9\%–11\% lead, validating the robustness and effectiveness of the model in extreme long-tail scenarios.

These improvements stem from LumiCRS's collaborative design across representation, optimization, and data. First, the Asymmetric Contrastive Prototype Learning constructs context-emotion-semantic prototypes for both head and tail movies, effectively mitigating representation instability caused by long-tail sparsity. Second, Adaptive Comprehensive Focal Loss (ACFL) dynamically suppresses head samples and amplifies the gradients of hard tail instances during training, guiding the model’s focus toward rare and difficult tail items. Lastly, a GPT-4o-based hierarchical dialogue augmentation module generates diverse conversational data specifically for tail movies and moderately enriches head items, reducing popularity bias at the data source. These three mechanisms reinforce each other, enabling LumiCRS to significantly improve long-tail recall, coverage, and diversity without sacrificing hit rate on popular items, thus demonstrating its efficacy in addressing long-tail challenges.

\begin{table*}[htbp]
\caption{\textbf{Long-tail performance on ReDial and INSPIRED.} 
Bold “*” = best with $p<0.05$;   \underline{value} = strongest baseline;   Gain = relative improvement over the strongest baseline.}
\label{tab:longtail}
\centering
\resizebox{\textwidth}{!}{
\begin{tabular}{c|ccc|cc|cc|ccc|cc|cc}
\toprule
 & \multicolumn{7}{c|}{\textbf{ReDial}} & \multicolumn{7}{c}{\textbf{INSPIRED}} \\
\cmidrule(lr){2-8} \cmidrule(lr){9-15}
\multirow{2}{*}{\textbf{Model}}
& \multicolumn{3}{c|}{\textbf{Tail-Recall}} 
& \multicolumn{2}{c|}{\textbf{Coverage}} 
& \multicolumn{2}{c|}{\textbf{ILD}}
& \multicolumn{3}{c|}{\textbf{Tail-Recall}} 
& \multicolumn{2}{c|}{\textbf{Coverage}} 
& \multicolumn{2}{c}{\textbf{ILD}} \\
\cmidrule(lr){2-4} \cmidrule(lr){5-6} \cmidrule(lr){7-8}
\cmidrule(lr){9-11} \cmidrule(lr){12-13} \cmidrule(lr){14-15}
& @1 & @10 & @50 & @10 & @50 & @10 & @50
& @1 & @10 & @50 & @10 & @50 & @10 & @50 \\
\midrule
Popularity & 0.008 & 0.025 & 0.041 & 0.111 & 0.158 & 0.094 & 0.110 & 0.008 & 0.023 & 0.038 & 0.115 & 0.164 & 0.088 & 0.103\\
TextCNN    & 0.009 & 0.027 & 0.045 & 0.120 & 0.172 & 0.104 & 0.122 & 0.008 & 0.025 & 0.042 & 0.125 & 0.179 & 0.098 & 0.115\\
BERT       & 0.014 & 0.041 & 0.068 & 0.165 & 0.235 & 0.123 & 0.145 & 0.012 & 0.036 & 0.060 & 0.169 & 0.242 & 0.118 & 0.139\\
ReDial     & 0.008 & 0.024 & 0.039 & 0.108 & 0.155 & 0.092 & 0.108 & 0.007 & 0.020 & 0.033 & 0.107 & 0.153 & 0.086 & 0.101\\
KBRD       & 0.015 & 0.044 & 0.074 & 0.206 & 0.294 & 0.133 & 0.157 & 0.012 & 0.037 & 0.062 & 0.194 & 0.277 & 0.125 & 0.147\\
KGSF       & 0.017 & 0.051 & 0.085 & 0.218 & 0.312 & 0.143 & 0.169 & 0.015 & 0.044 & 0.074 & 0.204 & 0.291 & 0.135 & 0.158\\
TREA       & 0.019 & 0.056 & 0.093 & 0.229 & 0.327 & 0.150 & 0.176 & 0.016 & 0.049 & 0.082 & 0.215 & 0.308 & 0.142 & 0.167\\
VRICR      & 0.018 & 0.053 & 0.089 & 0.223 & 0.319 & 0.145 & 0.171 & 0.015 & 0.046 & 0.076 & 0.209 & 0.296 & 0.138 & 0.162\\
UNICRS     & 0.020 & 0.059 & 0.099 & 0.235 & 0.336 & 0.152 & 0.179 & 0.017 & 0.053 & 0.088 & 0.226 & 0.331 & 0.150 & 0.176\\
DCRS      & \underline{0.021} & \underline{0.061} & \underline{0.102} & \underline{0.240} & \underline{0.342} & \underline{0.155} & \underline{0.182}
          & \underline{0.019} & \underline{0.056} & \underline{0.093} & \underline{0.234} & \underline{0.337} & \underline{0.181} & \underline{0.188} \\[-1pt]
MSCRS      & 0.017 & 0.052 & 0.087 & 0.236 & 0.346 & 0.162 & 0.194 & 0.016 & 0.050 & 0.084 & 0.234 & 0.334 & 0.157 & 0.188\\
DisenCRS   & 0.014 & 0.047 & 0.082 & 0.232 & 0.335 & 0.153 & 0.186 & 0.012 & 0.045 & 0.079 & 0.228 & 0.321 & 0.148 & 0.179\\
\rowcolor{gray!10}
\textbf{LumiCRS} & \textbf{0.024*} & \textbf{0.067*} & \textbf{0.114*} & \textbf{0.260*} & \textbf{0.370*} & \textbf{0.170*} & \textbf{0.200*}
                 & \textbf{0.022*} & \textbf{0.062*} & \textbf{0.105*} & \textbf{0.255*} & \textbf{0.367*} & \textbf{0.201*} & \textbf{0.209*}\\
\midrule
\textit{Gain (\%)} & 14.29\% & 9.84\% & 11.76\% & 8.33\% & 8.19\% & 9.68\% & 9.89\%
              & 15.79\% & 10.71\% & 12.90\% & 8.97\% & 8.90\% & 11.05\% & 11.17\%\\
\bottomrule
\end{tabular}}
\end{table*}

\subsection{Evaluation on Conversation Task (RQ2)}
\begin{table*}[tbp]
\centering
\setlength{\tabcolsep}{4pt}
\caption{\textbf{Automatic evaluation for the conversation task on ReDial and INSPIRED datasets.}  
Bold “*” = best with $p<0.05$;   \underline{value} = strongest baseline;   Gain = relative improvement over the strongest baseline.}
\label{tab:cov}
\resizebox{\textwidth}{!}{%
\begin{tabular}{c|cc|cc|ccc|cc|cc|ccc}
\toprule
            & \multicolumn{7}{c|}{\textbf{ReDial}} & \multicolumn{7}{c}{\textbf{INSPIRED}} \\[1pt]
\cmidrule(lr){2-8} \cmidrule(lr){9-15}
\multirow{2}{*}{\textbf{Model}}
            & \multicolumn{2}{c|}{\textbf{BLEU}} 
            & \multicolumn{2}{c|}{\textbf{ROUGE}} 
            & \multicolumn{3}{c|}{\textbf{DIST}} 
            & \multicolumn{2}{c|}{\textbf{BLEU}} 
            & \multicolumn{2}{c|}{\textbf{ROUGE}} 
            & \multicolumn{3}{c}{\textbf{DIST}} \\
\cmidrule(lr){2-3} \cmidrule(lr){4-5} \cmidrule(lr){6-8} \cmidrule(lr){9-10} \cmidrule(lr){11-12} \cmidrule(lr){13-15}
            & \textbf{-2} & \textbf{-3} & \textbf{-2} & \textbf{-L} & \textbf{-2} & \textbf{-3} & \textbf{-4}
            & \textbf{-2} & \textbf{-3} & \textbf{-2} & \textbf{-L} & \textbf{-2} & \textbf{-3} & \textbf{-4} \\ 
\midrule
DialogGPT   & 0.041 & 0.021 & 0.054 & 0.258 & 0.436 & 0.632 & 0.771 & 0.031 & 0.014 & 0.041 & 0.207 & 1.954 & 2.750 & 3.235 \\
GPT-2       & 0.031 & 0.013 & 0.041 & 0.244 & 0.405 & 0.603 & 0.757 & 0.026 & 0.011 & 0.034 & 0.212 & 2.119 & 3.084 & 3.643 \\
BART        & 0.024 & 0.011 & 0.031 & 0.229 & 0.432 & 0.615 & 0.705 & 0.018 & 0.008 & 0.025 & 0.208 & 1.920 & 2.501 & 2.670 \\
\midrule
ReDial      & 0.004 & 0.001 & 0.021 & 0.187 & 0.058 & 0.204 & 0.442 & 0.001 & 0.000 & 0.004 & 0.168 & 0.359 & 1.043 & 1.760 \\
KBRD        & 0.038 & 0.018 & 0.047 & 0.237 & 0.070 & 0.288 & 0.488 & 0.021 & 0.007 & 0.029 & 0.218 & 0.416 & 1.375 & 2.320 \\
KGSF        & 0.030 & 0.012 & 0.039 & 0.244 & 0.061 & 0.278 & 0.515 & 0.023 & 0.007 & 0.031 & 0.228 & 0.418 & 1.496 & 2.790 \\
VRICR       & 0.021 & 0.008 & 0.027 & 0.137 & 0.107 & 0.286 & 0.471 & 0.011 & 0.001 & 0.025 & 0.187 & 0.853 & 1.801 & 2.827 \\
TREA        & 0.022 & 0.008 & 0.039 & 0.175 & 0.242 & 0.615 & 1.176 & 0.013 & 0.002 & 0.027 & 0.195 & 0.958 & 2.565 & 3.411 \\
COLA        & 0.026 & 0.012 & --    & --    & 0.387 & 0.528 & 0.625 & --    & --    & --    & --    & --    & --    & --    \\
UNICRS      & 0.045 & 0.021 & 0.058 & 0.285 & 0.433 & 0.748 & 1.003 & 0.022 & 0.009 & 0.029 & 0.212 & 2.686 & 4.343 & 5.520 \\
DCRS        & 0.048 & 0.024 & 0.063 & 0.285 & 0.779 & 1.173 & 1.386 & 0.033 & 0.014 & 0.045 & 0.229 & 3.950 & 5.729 & 6.233 \\
DisenCRS    & 0.050 & 0.027 & 0.065 & 0.286 & 0.784 & 1.212 & 1.411 & 0.034 & 0.017 & 0.048 & 0.232 & 4.014 & \underline{6.318} & \underline{6.623} \\
MSCRS       & \underline{0.054} & \underline{0.027} & \underline{0.070} & \underline{0.294} & \underline{0.784} & \underline{1.332} & \underline{1.553}
            & \underline{0.040} & \underline{0.019} & \underline{0.052} & \underline{0.235} & \underline{4.197} & 5.983 & 6.556 \\
\rowcolor{gray!10}
\textbf{LumiCRS} 
            & \textbf{0.058*} & \textbf{0.030*} & \textbf{0.074*} & \textbf{0.303*} & \textbf{0.820*} & \textbf{1.450*} & \textbf{1.710*}
            & \textbf{0.044*} & \textbf{0.021*} & \textbf{0.056*} & \textbf{0.242*} & \textbf{4.400*} & \textbf{6.900*} & \textbf{7.400*} \\
\midrule
\textit{Gain (\%)} 
            & 7.41\% & 11.11\% & 5.71\% & 3.06\% & 4.59\% & 8.86\% & 10.11\%
            & 10.00\% & 10.53\% & 7.69\% & 2.98\% & 4.84\% & 9.21\% & 11.73\% \\
\bottomrule
\end{tabular}}
\end{table*}
\begin{table}[ht]
\centering
\caption{\textbf{Subjective dialogue quality (1–5).}  
Bold “*” = best with $p<0.05$;   \underline{value} = strongest baseline;   Gain = relative improvement over the strongest baseline.}
\label{tab:subjective}
\begin{tabular}{lccccc}
\toprule
Model & Fluency & Info & Persua & Human & TailRel \\
\midrule
KBRD       & 2.92 & 2.57 & 2.44 & 2.71 & 2.10 \\
KGSF       & 3.06 & 2.65 & 2.52 & 2.85 & 2.28 \\
VRICR      & 2.97 & 2.77 & 2.63 & 2.76 & 2.31 \\
UniCRS     & 3.31 & 3.13 & 2.92 & 3.02 & 2.56 \\
DCRS       & 3.40 & \underline{3.25} & 3.01 & 3.15 & 2.74 \\
\underline{MSCRS} & 3.42 & 3.24 & \underline{3.03} & 3.18 & 2.78 \\
DisenCRS   & \underline{3.45} & 3.22 & 2.98 & \underline{3.20} & \underline{2.81} \\
\rowcolor{gray!10}
\textbf{LumiCRS}   & \textbf{3.65*} & \textbf{3.45*} & \textbf{3.30*} & \textbf{3.45*} & \textbf{3.20*} \\
\midrule
\textit{Gain (\%)} & 5.8\%          & 6.2\%          & 8.9\%          & 7.8\%          & 13.9\% \\

\bottomrule
\end{tabular}
\end{table}
As shown in Table~\ref{tab:cov} (automatic metrics) and Table~\ref{tab:subjective} (subjective metrics), LumiCRS achieves state-of-the-art performance on both ReDial and INSPIRED; the two evaluation protocols corroborate each other across all dimensions.

On ReDial, LumiCRS attains BLEU-2/3 of 0.058 and 0.030, surpassing the strongest baseline (MSCRS) by 7.4\% and 11.1\%, respectively; the gains further enlarge to 10.0\% and 10.5\% on the sparser INSPIRED dataset.  
Diversity improvements are even more pronounced: DIST-4 on ReDial rises to 1.710, a 10.1\% relative gain, while DIST-2/3/4 on INSPIRED increase by 4.8\%, 9.2\% and 11.7\%, indicating richer and less repetitive responses and stronger utilisation of long-tail vocabulary.

Crowd annotators rate LumiCRS highest on all five dimensions.  
Compared with the best baseline (DisenCRS), Fluency and Human-likeness improve by 5.8\% and 7.8\%, validating the naturalness of the generated utterances; Informativeness and Persuasiveness rise by 6.2\% and 8.9\%, reflecting more substantial and guiding content.  
Crucially, the TailRel score reaches 3.20 (a 13.9\% lift), echoing the automatic diversity gains and confirming LumiCRS’s ability to surface relevant long-tail information.

LumiCRS weaves three complementary layers to tame the long-tail problem: (1) Prototype Learning for Long-Tail Recommendation builds context-, affect-, and semantic-aware prototypes for body and tail movies, yielding stable representations under extreme sparsity; (2) Adaptive Comprehensive Focal Loss (ACFL) dynamically down-weights over-represented head samples while amplifying hard tail instances, suppressing popularity bias and repetition; and (3) GPT-4o-Based Prototype Dialogue Data Augmentation injects diverse, quality-controlled utterances around tail prototypes to enrich supervision. Acting in concert, these modules push BLEU and ROUGE to new highs and boost DIST-k and TailRel by ≥ 10\%, delivering simultaneous gains in response quality and long-tail coverage.

\subsection{Ablation Study (RQ3)}
\begin{figure}
    \centering
    \includegraphics[width=1\linewidth]{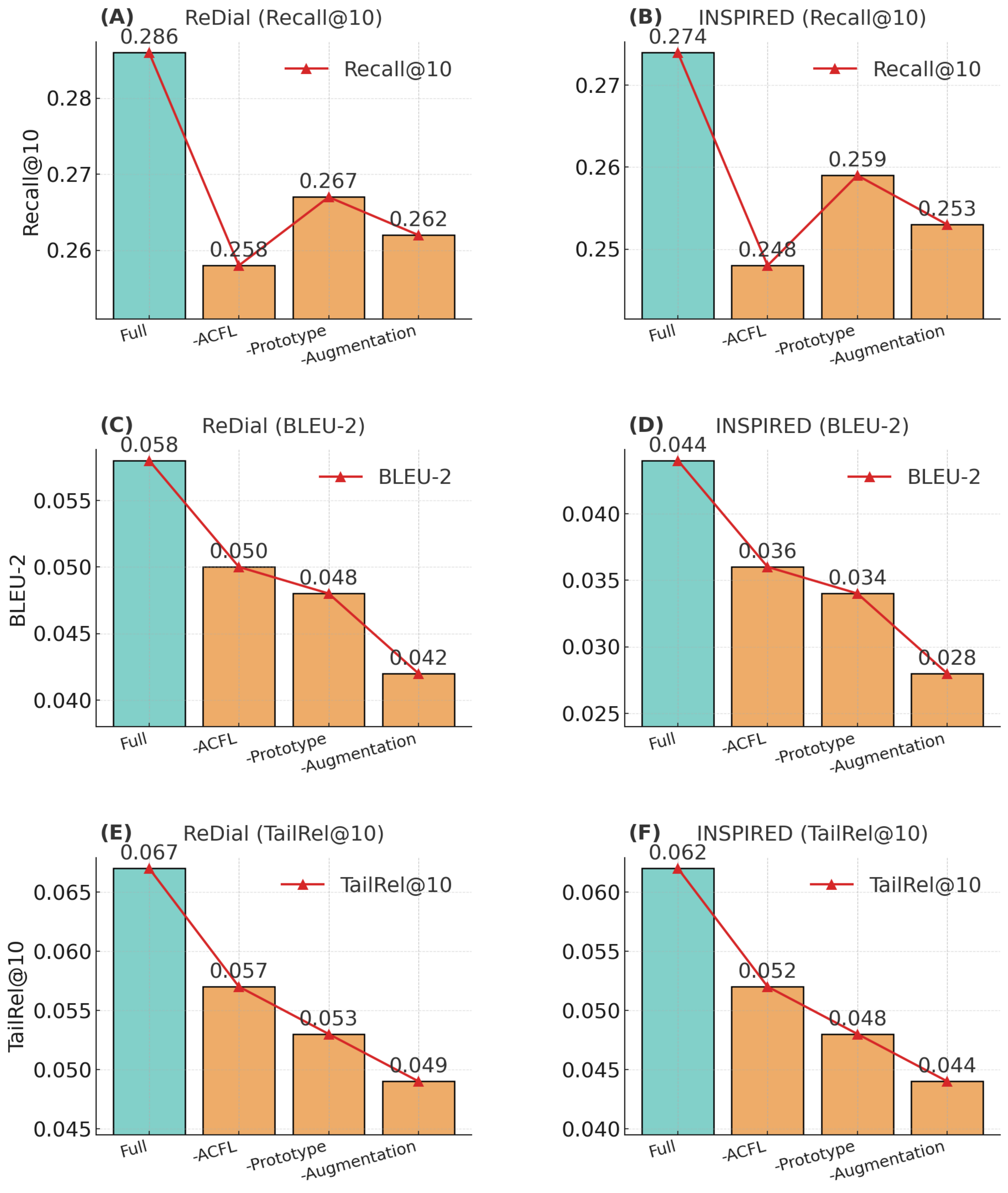}
    \caption{\textbf{Ablation study of LumiCRS on ReDial and INSPIRED.}
Removing ACFL, prototype learning, or GPT-4o dialogue augmentation individually causes notable drops in Recall@10, BLEU-2, and TailRel@10, confirming that the three modules are complementary and jointly critical for accuracy, fluency, and long-tail coverage.
} 
    \label{Ablation Study}
\end{figure}
To quantify the individual contribution of each core component in LumiCRS, we conducted ablation experiments on the ReDial and INSPIRED benchmarks. Four model variants were compared:
Full – the complete LumiCRS framework;
-ACFL – the Adaptive Comprehensive Focal Loss (ACFL) is entirely replaced by standard Cross-Entropy (CE) while keeping the loss structure intact;
-Prototype – the prototype-learning module is removed;
-Augmentation – the GPT-4o-driven multi-turn dialogue augmentation is disabled.
All other hyper-parameters and training protocols remain unchanged. We report Recall@10 (overall recommendation accuracy), BLEU-2 (response fluency), and TailRel@10 (long-tail hit rate). Results are visualised in Figures\ref{Ablation Study}.

Overall recommendation accuracy (Recall@10)：
In both datasets, the Full model consistently achieves the best Recall@10 ($\approx 0.28$)
. Replacement of ACFL cement with CE leads to a drop - about 9 - 10\%, underscoring the role of ACFL the role of ACFL in mitigating popularity bias and mining hard examples. Removing the prototype module or dialogue augmentation also hurts performance, though to a lesser extent ($\approx 5\hyp8\%$).

Response generation quality (BLEU-2):
Figures\ref{Ablation Study}c and \ref{Ablation Study}d show that BLEU-2 is most sensitive to data augmentation: disabling GPT-4o augmentation degrades BLEU-2 by 28 – 36\%, severely harming lexical diversity and semantic coherence. Eliminating prototypes causes a drop of 17 - 23\%, while substituting ACFL with CE still reduces BLEU-2 by approximately 15\%, indicating that a stable recommendation core indirectly benefits language generation.

Long-tail recommendation ability (TailRel@10):
Figures\ref{Ablation Study}e and \ref{Ablation Study}f reveal a similar pattern for long-tail coverage. Removing augmentation slashes TailRel@10 by $\approx30\%$, confirming that diverse synthetic dialogues effectively alleviate tail-data sparsity. Replacement of ACFLs causes a 15 - 16\% decline, and removal of prototypes cuts another 20\%+, demonstrating their importance in strengthening representations of the middle and tail segments.

Based on the above observations, three key conclusions can be drawn:
First, the loss layer (ACFL), representation layer (prototype learning), and data layer (dialogue augmentation) form an interdependent and mutually compensatory triad. The absence of any single component leads to simultaneous and significant degradation across all three main objectives—recommendation accuracy, response quality, and long-tail coverage.
Second, even merely replacing ACFL with CE—rather than removing it—results in double-digit performance drops, highlighting the unique strength of the Adaptive Comprehensive Focal Loss in handling highly imbalanced scenarios.
Third, dialogue augmentation plays a particularly critical role in enhancing fluency and long-tail coverage, while ACFL serves as the backbone for maintaining overall Recall and emphasizing tail gradients. Prototype learning, meanwhile, provides indispensable support for stabilizing cross-domain semantic representations.
\subsection{ACFL for Bias and Overfitting (RQ4)}

To evaluate whether our proposed Adaptive Comprehensive Focal Loss (ACFL) effectively mitigates popularity bias and training overfitting, we compare it with eight representative loss functions: Cross Entropy (CE), Focal Loss, Adaptive Logit Adjustment (ALA), Enhanced Adaptive Focal Loss (EAFL), Asymmetric Loss with Padé Approximation (ALPA), Cost-Sensitive Loss (CSL), Distributionally Robust Loss (DR-Loss), and Adaptive Hierarchical Loss (AHL). All experiments are conducted under the same LumiCRS model configuration and training regime, varying only the loss function to ensure fair comparisons.

Table~\ref{tab:loss_comparison} reports results on ReDial across three key metrics that align with our core motivation: overall accuracy (Recall@10), long-tail item coverage (TailRecall@10), and popularity-weighted precision (PWP). ACFL achieves the best performance on all three metrics: 0.286 Recall@10, 0.067 TailRecall@10, and the lowest 0.045 PWP, outperforming even the second-best DR-Loss (0.265, 0.061, 0.052 respectively). These gains confirm that ACFL maintains overall accuracy while explicitly improving long-tail representation and reducing head-class bias.

\begin{figure}[t]
    \centering
    \includegraphics[width=1\linewidth]{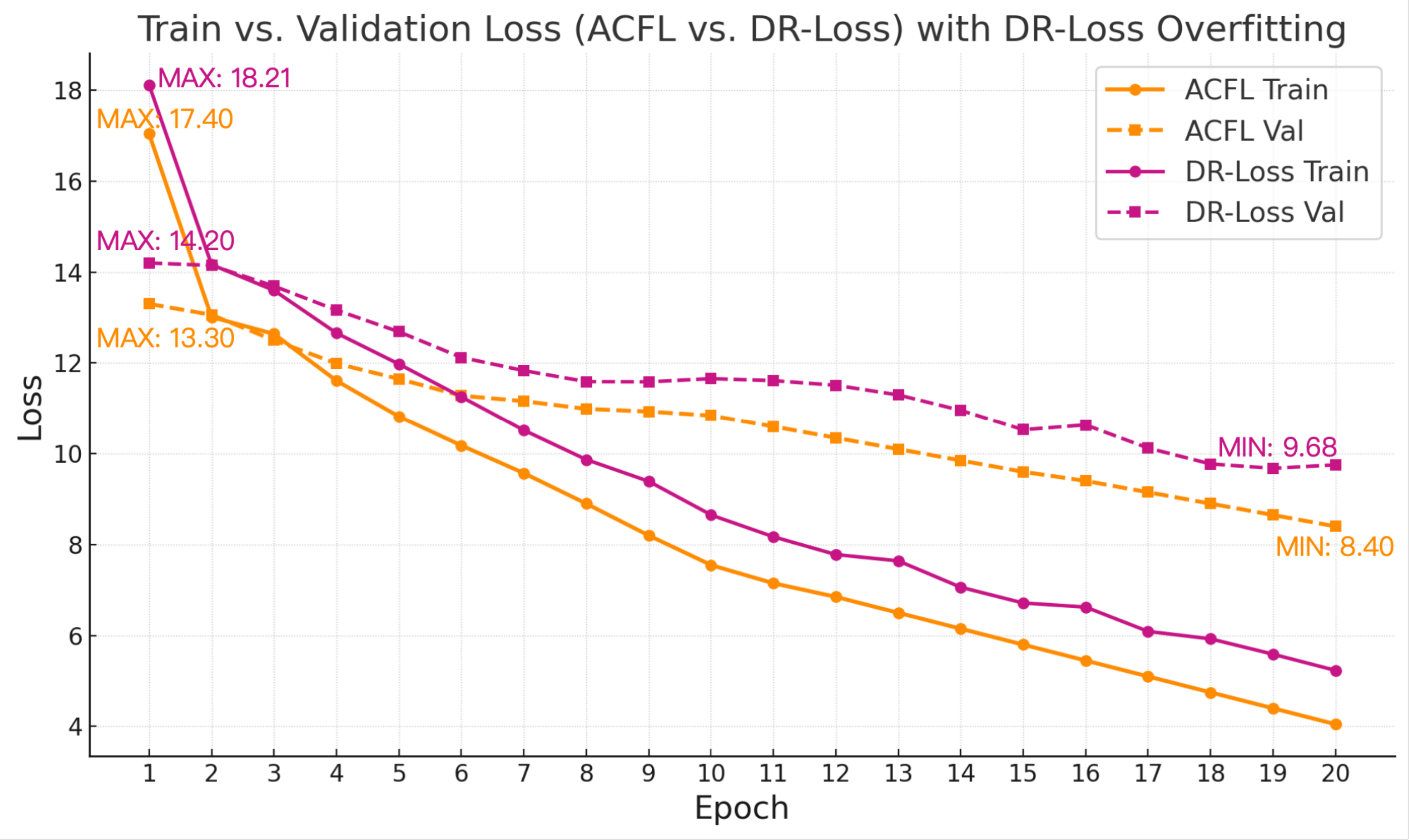} 
    \caption{\textbf{Training vs. validation loss comparison between ACFL and DR-Loss on ReDial.} ACFL maintains smooth convergence while DR-Loss shows late-stage overfitting.}
    \label{fig:loss_curve_acfl_vs_drloss}
\end{figure}

Additionally, we visualize the training and validation loss dynamics in Figure~\ref{fig:loss_curve_acfl_vs_drloss}. Compared to DR-Loss, which suffers from validation loss stagnation and noticeable overfitting in the later epochs, ACFL continues to optimize both training and validation loss smoothly, demonstrating its stability and superior generalization. This aligns with our motivation that a dynamic, curriculum-based modulation of focal weight can guide models away from overfitting and head-class overconfidence, thereby encouraging balanced learning across item frequency spectra.

\begin{table}[t]
\centering
\caption{\textbf{Comparison of loss functions on ReDial.} ACFL significantly improves both overall accuracy and long-tail item coverage while reducing popularity bias.}
\label{tab:loss_comparison}
\begin{tabular}{lccc}
\toprule
\textbf{Loss Function} & \textbf{Recall@10} ↑ & \textbf{TailRecall@10} ↑ & \textbf{PWP} ↓ \\
\midrule
Cross Entropy      & 0.232 & 0.031 & 0.076 \\
Focal Loss             & 0.236 & 0.036 & 0.072 \\
ALA (2022)                    & 0.241 & 0.039 & 0.068 \\
EAFL (2024)                   & 0.243 & 0.042 & 0.066 \\
ALPA (2024)                   & 0.240 & 0.040 & 0.067 \\
CSL (2025)                    & \underline{0.265} & 0.058 & 0.052 \\
DR-Loss (2024)                & 0.258 & \underline{0.061} & \underline{0.052} \\
AHL (2025)                    & 0.257 & 0.059 & 0.053 \\
\rowcolor{gray!10}
\textbf{ACFL (Ours)}   & \textbf{0.286} & \textbf{0.067} & \textbf{0.045} \\
\midrule

\textit{Gain\%} & 7.92\% & 9.84\% & -13.46\% \\

\bottomrule
\end{tabular}
\end{table}

\subsection{Prototype Learning for Robust Mid Frequency Representation (RQ5)}

To investigate the effectiveness of the proposed Prototype Learning (PL) module in improving the semantic representation of mid-frequency movies, we conduct a series of ablation studies on the ReDial dataset. Specifically, we evaluate the following three variants:
(i) w/o Prototype: The PL module is completely removed.
(ii) + Random Prototype: The learnable prototypes are replaced with fixed, randomly initialized vectors.
(iii) + KMeans Init: Prototypes are initialized with cluster centers from KMeans but frozen during training.
    
As shown in Table~\ref{tab:proto_ablation_redial}, removing or weakening the prototype mechanism consistently leads to performance drops across all metrics. The \textit{w/o Prototype} variant performs worst, highlighting the necessity of structured, learnable prototypes in modeling complex user-item semantics. Randomly initialized prototypes offer only marginal gains, suggesting that static embeddings are insufficient. The \textit{+ KMeans Init} variant performs slightly better than random, but still underperforms the full model, indicating that dynamic, end-to-end training is crucial for exploiting the full potential of prototypes.

\begin{figure}
    \centering
    \includegraphics[width=1\linewidth]{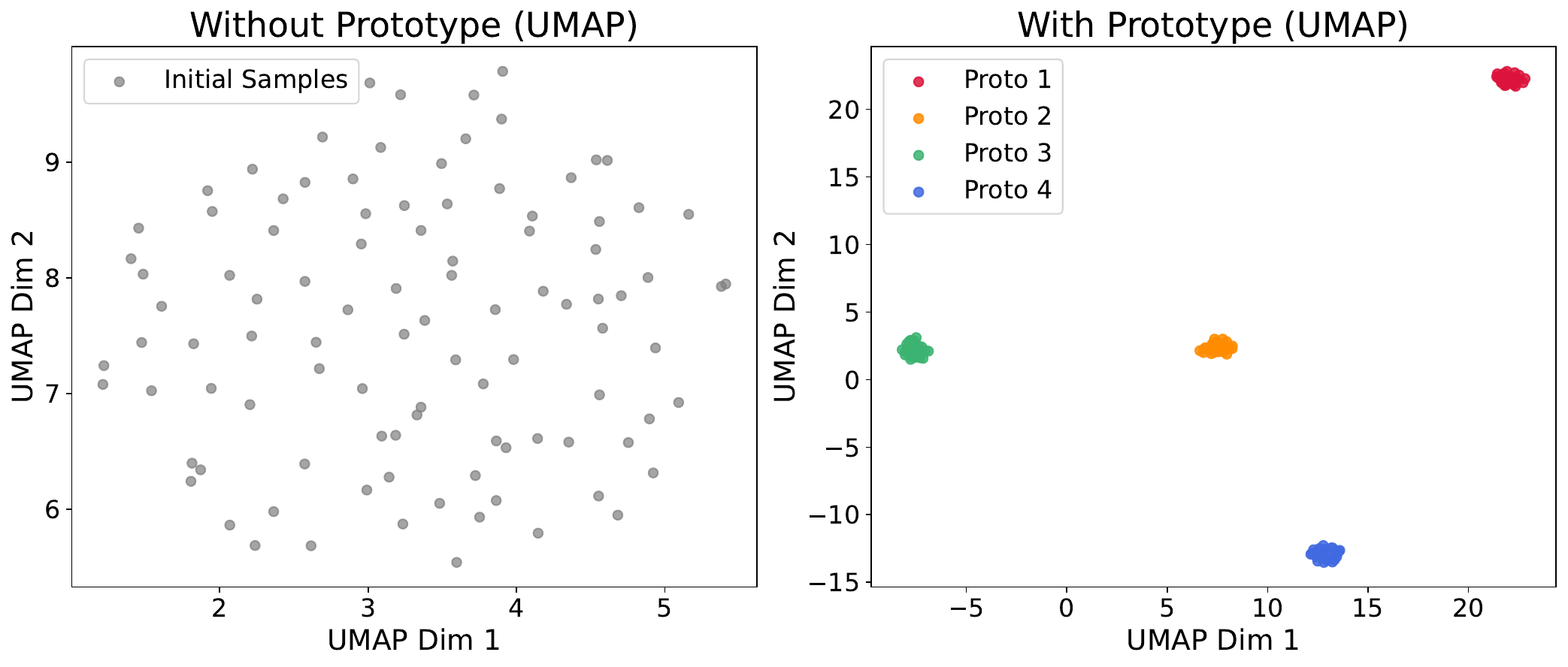}
    \caption{\textbf{UMAP visualization of mid-frequency movie embeddings under two settings.}
  (a) \textit{w/o Prototype}—points are scattered and loosely organized, indicating unstable and
  overlapping semantic representations; 
  (b) \textit{Full}—with prototype learning enabled, embeddings form compact and well-separated
  clusters, demonstrating improved semantic coherence and discriminability for mid-frequency items.}
    \label{fig:umap_proto}
\end{figure}

In addition to quantitative analysis, we visualize the embedding structures of mid-frequency movies using UMAP in Figure~\ref{fig:umap_proto}. Compared to the scattered and disorganized distribution in the \textit{w/o Prototype} case, the full model exhibits more compact and clearly clustered embeddings, indicating stronger semantic coherence and improved robustness.

These results confirm that our prototype learning strategy plays a critical role in modeling items in the mid frequency range, addressing one of the key motivations of this work by improving the quality of the representation and generalization.

\begin{table}[t]
\centering
\caption{Ablation results on mid-frequency movies (ReDial).}
\label{tab:proto_ablation_redial}
\begin{tabular}{lccc}
\toprule
\textbf{Model Variant} & \textbf{Recall@10} & \textbf{TailRecall@10}  & \textbf{NDCG@10}  \\
\midrule
-Prototype            & 0.219 & 0.044 & 0.163 \\
Random   & 0.225 & 0.046 & 0.167 \\
KMeans           & 0.230 & 0.049 & 0.171 \\
\rowcolor{gray!10}
\textbf{Full} & \textbf{0.238} & \textbf{0.051} & \textbf{0.174} \\
\bottomrule
\end{tabular}
\end{table}
\subsection{Prototype-Driven Augmentation for Long-Tail Enhancement (RQ6)}
To assess whether prototype-driven augmentation (PDA) can enhance long-tail and cold-start performance, we conduct an ablation study comparing several augmentation strategies: (1) w/o Augmentation, where the model is trained without any additional contextual augmentation; (2) + Random Augmentation, which incorporates randomly sampled dialogues; (3) + Static Template, using handcrafted sentence templates; and (4) w/ PDA, where semantic prototypes dynamically guide the generation of diverse, contextually relevant augmentations.

We evaluate the models using three key metrics: overall accuracy (Recall@10), long-tail item coverage (TailRecall@10), and generation diversity (Distinct-2). As shown in Table~\ref{tab:augmentation_comparison}, the PDA-augmented model significantly outperforms the others across all three dimensions. Notably, it achieves the highest Recall@10 (0.286), the best tail coverage (0.067), and the most diverse generation (0.820 Dist-2) on the ReDial dataset.

\begin{table}[t]
\centering
\caption{\textbf{Performance of different augmentation strategies on ReDial.} Prototype-driven augmentation (PDA) achieves the best balance between accuracy, long-tail coverage, and generation diversity.}
\label{tab:augmentation_comparison}
\begin{tabular}{lccc}
\toprule
\textbf{Strategy} & \textbf{Recall@10}& \textbf{TailRecall@10}& \textbf{Distinct-2}\\
\midrule
w/o Aug      & 0.261 & 0.042 & 0.742 \\
+ Random Aug & 0.266 & 0.048 & 0.768 \\
+ Static Tem      & 0.272 & 0.056 & 0.794 \\
\rowcolor{gray!10}
\textbf{w/ PDA (Full)}         & \textbf{0.286} & \textbf{0.067} & \textbf{0.820} \\
\bottomrule
\end{tabular}
\end{table}
\begin{figure}
    \centering
    \includegraphics[width=1\linewidth]{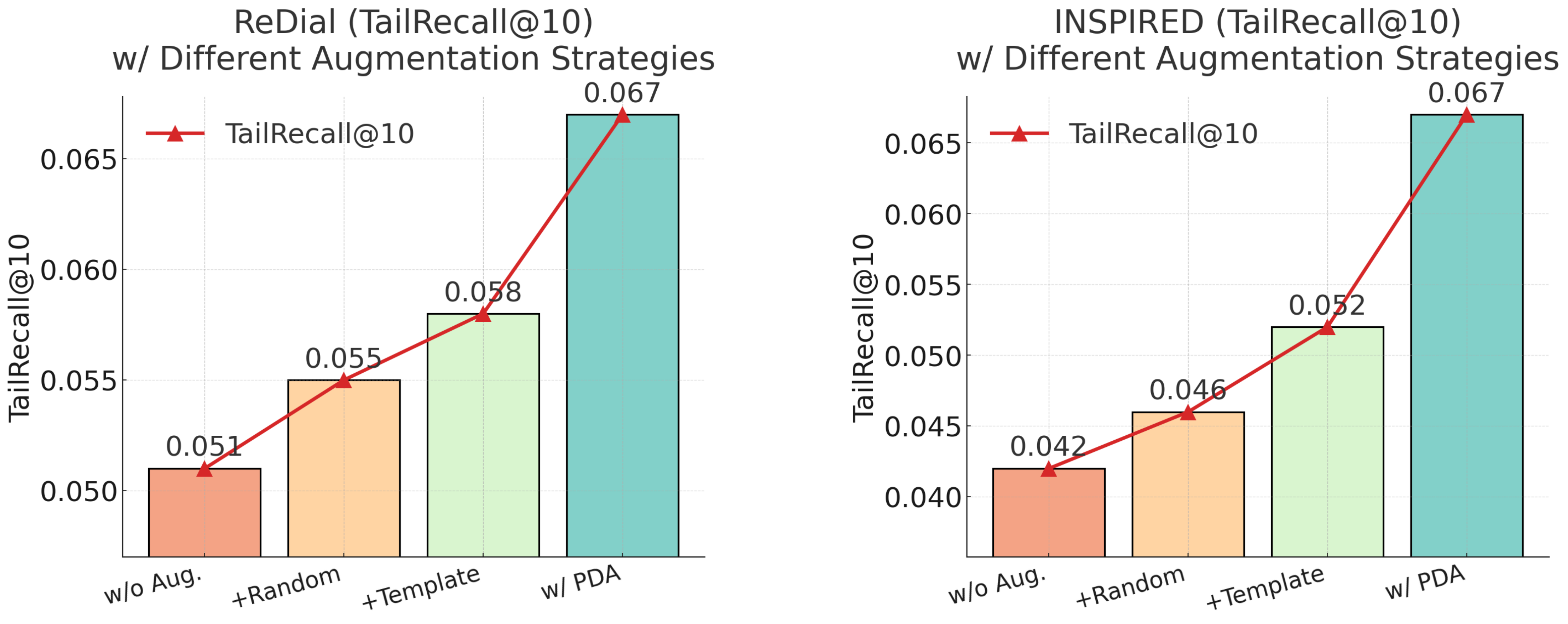}
    \caption{\textbf{TailRecall@10 under different augmentation strategies on ReDialand INSPIRED.}
Prototype-driven augmentation (w/ PDA) consistently achieves the highest long-tail recall, outperforming models trained without augmentation (w/o Aug.), with random augmentation (+ Random), or with static templates (+ Template).}
    \label{tailrecall}
\end{figure}

To further validate the effectiveness of PDA on long-tail recommendation, we visualize TailRecall@10 across different strategies on both ReDial and INSPIRED datasets. As illustrated in Figure~\ref{tailrecall}(A) and Figure~~\ref{tailrecall}(B), PDA consistently yields the highest tail recall, demonstrating its strong generalization ability in mitigating popularity bias and enhancing the representation of under-represented items.

\subsection{Hyper parameter Study (RQ7)}
\begin{figure}
    \centering
    \includegraphics[width=1\linewidth]{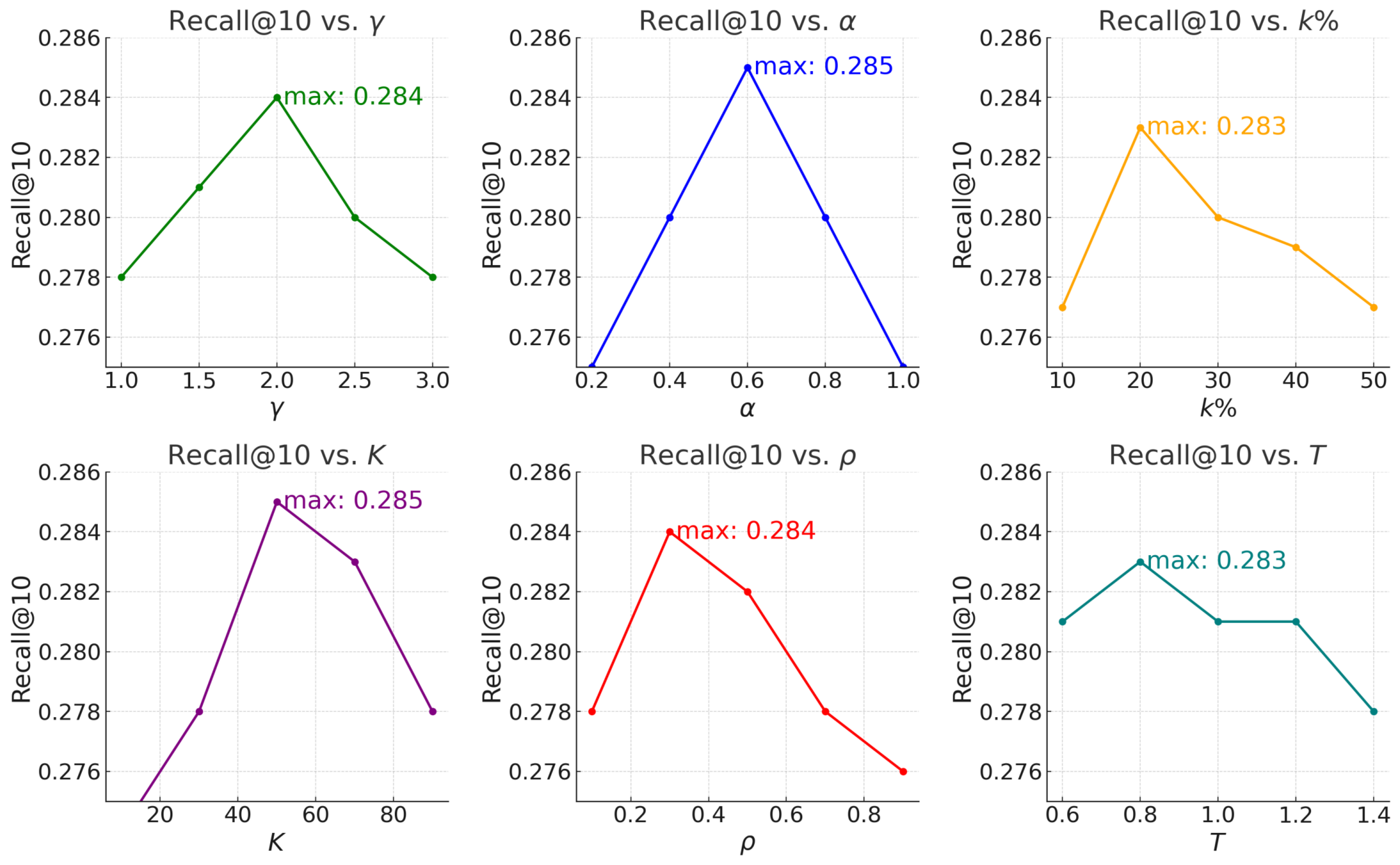}
    \caption{One-Factor Hyperparameter Sensitivity Analysis on ReDial (Recall@10).
}
    \label{ccs-1}
\end{figure}
\begin{figure}
    \centering
    \includegraphics[width=1\linewidth]{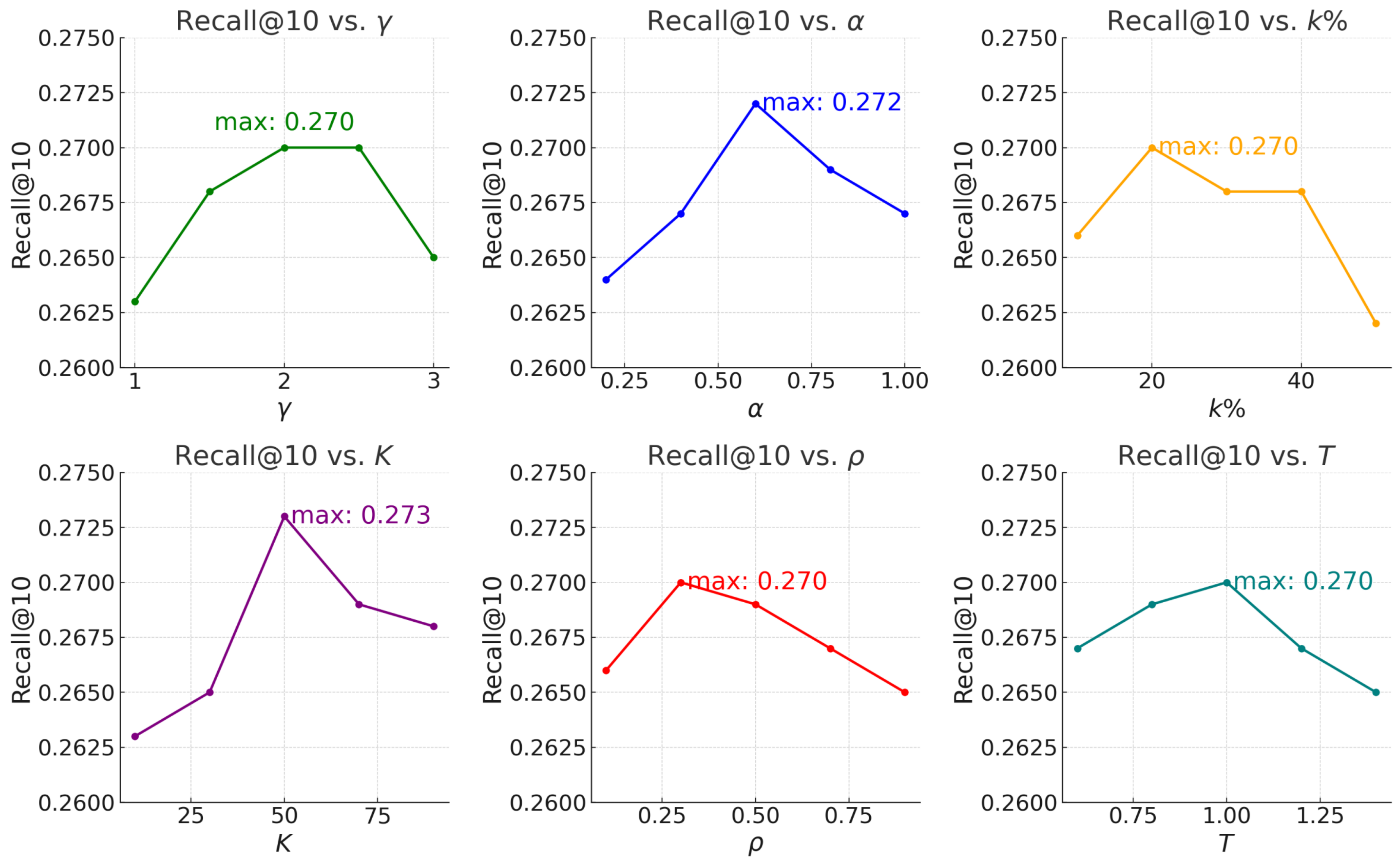}
    \caption{One-Factor Hyperparameter Sensitivity Analysis on INSPIRED (Recall@10).}
    \label{ccs-2}
\end{figure}
\begin{figure}
    \centering
    \includegraphics[width=1\linewidth]{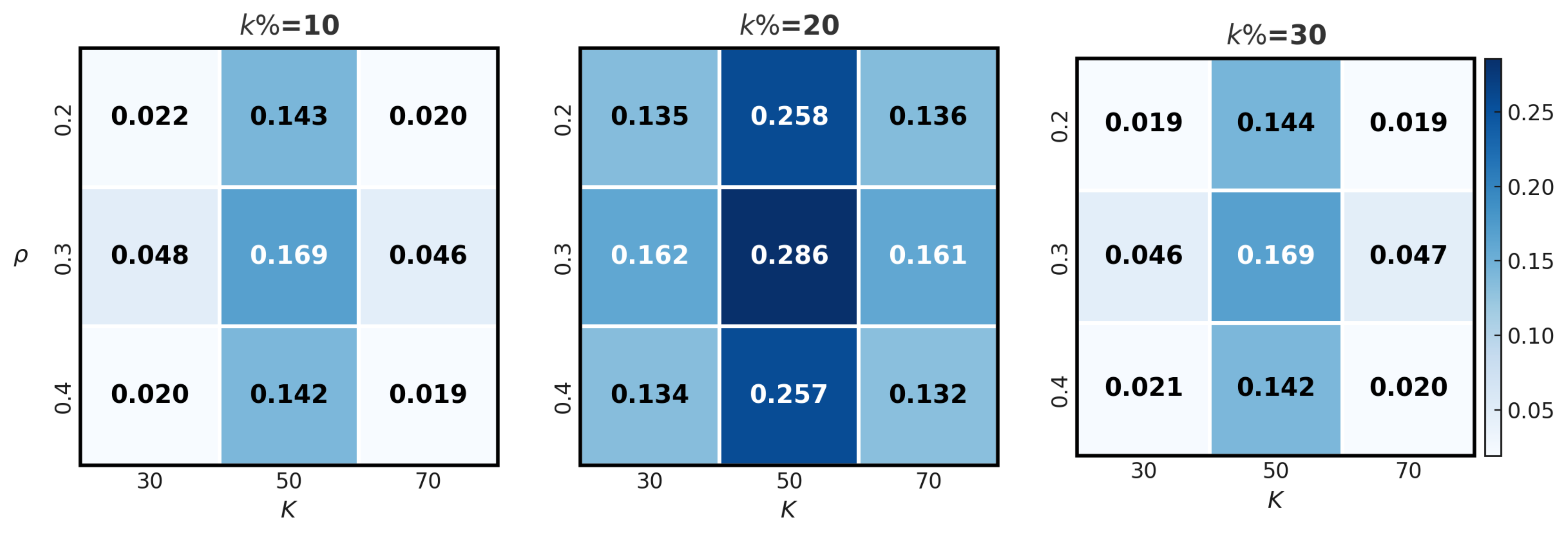}
    \caption{Hyperparameter Interaction Analysis on ReDial (Recall@10).}
    \label{ccs-3}
\end{figure}
\begin{figure}
    \centering
    \includegraphics[width=1\linewidth]{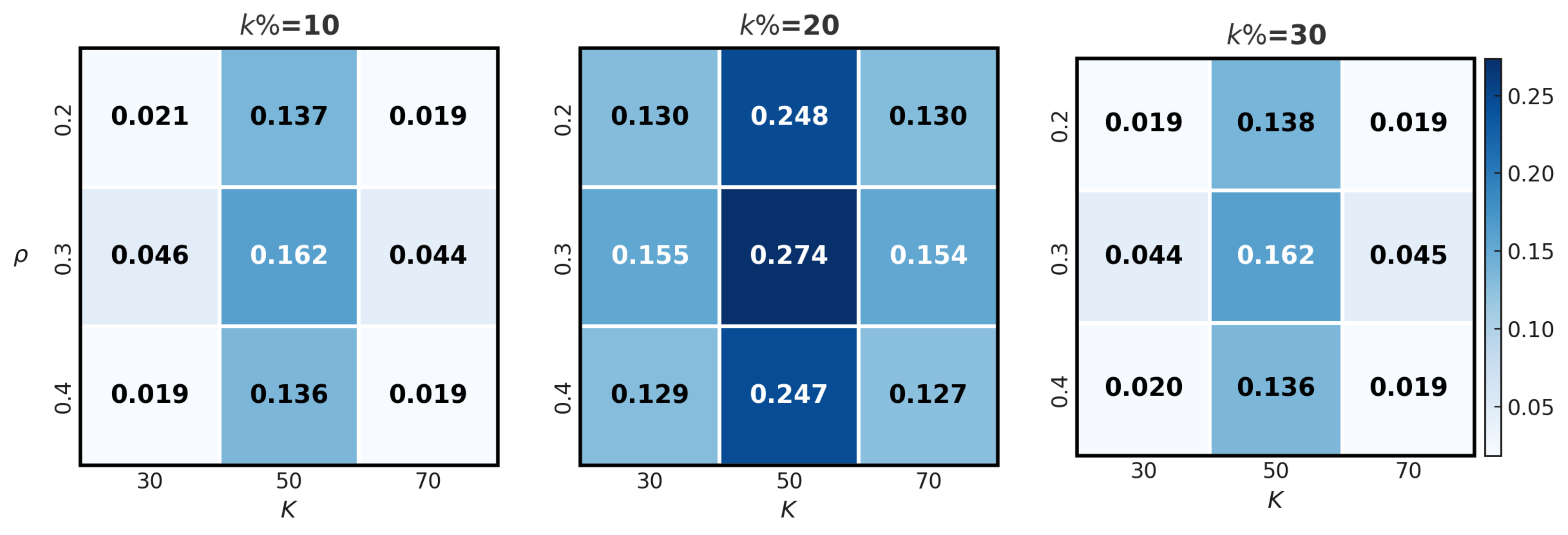}
    \caption{Hyperparameter Interaction Analysis on INSPIRED (Recall@10).}
    \label{ccs-4}
\end{figure}
To investigate the sensitivity of LumiCRS to its design choices, we evaluate six representative hyperparameters across three core components: (1) the optimization layer (ACFL), including the focusing parameter $\gamma$, the class-balancing factor $\alpha$, and the top-k hard example ratio $k\%$; (2) the representation layer, where $K$ denotes the number of prototypes; and (3) the data augmentation layer, involving the augmentation ratio $\rho$ and the augmentation temperature $T$. These parameters directly influence model generalization, especially for long-tail recommendation. Other parameters such as semantic similarity weight, emotion similarity weight, and interaction coefficient are not tuned here as preliminary experiments show minimal impact (less than 1.5\% performance variation), and stable values are adopted from previous studies.

We first conduct a one-factor-at-a-time sensitivity analysis (Fig.~\ref{ccs-1} and Fig.~\ref{ccs-2}) to observe how each parameter individually affects Recall@10 on ReDial and INSPIRED. Most parameters exhibit a peak in the mid-range, indicating the model’s general robustness within a certain interval. The optimal values are observed at $\gamma=2.5$, $\alpha=0.6$, $k\%=25$, $K=50$, $\rho=0.3$, and $T=0.7$, which yield the highest Recall@10 in both datasets, confirming their critical role.

Next, we select three hyperparameters with the most significant influence—$k\%$, $K$, and $\rho$—for a $3\times3\times3$ factorial interaction experiment. The heatmaps in Fig.~\ref{ccs-3} and Fig.~\ref{ccs-4} visualize the joint effects. Results indicate clear interaction effects: a moderate $k\%$ ensures meaningful long-tail signals, an appropriate number of prototypes $K$ maintains representation capacity without overfitting, and a balanced augmentation ratio $\rho$ enriches data without introducing excessive noise. The best configuration, $k\%=20$, $K=50$, and $\rho=0.3$, consistently outperforms others, reinforcing the robustness of the parameter setting found in the single-variable analysis.
\section{Case Studies} 
\begin{table}[t]
\small
\centering
\setlength\tabcolsep{6.0pt}
\caption{A sampled case extracted from the ReDial dataset. Text in a \sethlcolor{lime}\hl{green box} represents expressing emotions.}
\label{tab:case_study_bigdaddy}
\begin{tabularx}{\linewidth}{lX}
  \toprule
  \multirow{2}{*}{\rotatebox[origin=c]{90}{\textbf{Hist.}}}
   & \textbf{User}: I really liked \textit{Big Daddy}. Can you recommend another good comedy like that? \\
   \midrule
  {\multirow{16}{*}{\rotatebox[origin=c]{90}{\textbf{Response}}}}

  &{\textbf{Standard response}}: How about \textit{Grown Ups}? It's another Adam Sandler comedy. \\
  & \emph{\textbf{Fluency:} 3.0;\enspace\textbf{Info:} 1.5;\enspace\textbf{Persua:} 1.6;\enspace\textbf{Human:} 1.8;\enspace\textbf{TailRel:} 0.8};\\
  \cmidrule(r){2-2}

  &{\textbf{GPT4o}}: If you enjoyed \textit{Big Daddy}, you might also like \textit{Happy Gilmore} or \textit{Billy Madison}. They both feature Adam Sandler’s unique brand of humor — silly but heartfelt. \textit{Happy Gilmore} is about a failed hockey player who becomes an unlikely golf star, while \textit{Billy Madison} follows a man-child redoing all 12 school grades to inherit his father's company. \sethlcolor{lime}\hl{They’re both funny in that goofy, nostalgic way and could really hit the spot if you’re in the mood for a good laugh!} \\
  & \emph{\textbf{Fluency:} 3.9;\enspace\textbf{Info:} 3.1;\enspace\textbf{Persua:} 3.0;\enspace\textbf{Human:} 3.2;\enspace\textbf{TailRel:} 1.5};\\
  \cmidrule(r){2-2}

  &{\textbf{LLaMA2-Chat}}: Maybe you could check out \textit{Happy Gilmore} — it’s another comedy classic starring Adam Sandler. The story revolves around an aggressive hockey player who tries his luck in golf to save his grandmother’s house. It’s packed with over-the-top humor and bizarre sports antics. \sethlcolor{lime}\hl{If you liked the mix of sweetness and silliness in \textit{Big Daddy}, this one should hit a similar vibe.} Sandler’s character is once again outrageous but oddly lovable, and the film delivers some solid laughs throughout.\\
  & \emph{\textbf{Fluency:} 3.8;\enspace\textbf{Info:} 2.6;\enspace\textbf{Persua:} 2.7;\enspace\textbf{Human:} 3.3;\enspace\textbf{TailRel:} 1.4};\\
  \cmidrule(r){2-2}

  &{\textbf{LumiCRS[LLaMA2-Chat]}}: \sethlcolor{lime}\hl{So glad you liked \textit{Big Daddy} — it’s a classic!} Based on your taste, I’d recommend \textit{\textbf{The Way Way Back}}, a heartfelt and funny coming-of-age comedy. It’s subtle but impactful. \sethlcolor{lime}\hl{You’ll probably relate to the emotional beats and enjoy the characters.} Another pick would be \textit{\textbf{Sing Street}}, which blends great music with teen awkwardness in a sweet and uplifting way.\\
  & \emph{\textbf{Fluency:} 4.3;\enspace\textbf{Info:} 4.0;\enspace\textbf{Persua:} 3.9;\enspace\textbf{Human:} 4.1;\enspace\textbf{TailRel:} 3.8};\\

  \bottomrule
\end{tabularx}
\end{table}

Table~\ref{tab:case_study_bigdaddy} presents a case from the \textsc{ReDial} corpus in which the user states that they “really liked \textit{Big Daddy}” and requests “another good comedy like that.”  
Four single-turn responses are compared: the platform’s default reply, \textsc{GPT-4o}, \textsc{LLaMA2-Chat}, and our LumiCRS.  
Each answer is assessed on five 0–5 scales—\textit{Fluency}, \textit{Info}, \textit{Persua}, \textit{Human}, and \textit{TailRel}—shown in italics beneath the text.  

The baseline systems concentrate on well-known Adam Sandler titles such as \textit{Happy Gilmore} and \textit{Billy Madison}.  
Although \textsc{GPT-4o} and \textsc{LLaMA2-Chat} provide brief plot summaries and light emotive language, their long-tail relevance remains low ($TailRel \leq 1.5$).  
The default reply is even more limited, offering only \textit{Grown Ups} and scoring poorly on information and human-likeness.

By contrast, LumiCRS opens with an empathic remark (“So glad you liked \textit{Big Daddy} —it’s a classic!”) and recommends two thematically aligned yet much rarer films: the coming-of-age comedy The Way Way Back and the musical dramedy Sing Street.  
This response attains the highest scores in all five dimensions (Fluency 4.3, Info 4.0, Persua 3.9, Human 4.1, TailRel 3.8), demonstrating greater linguistic naturalness, richer justification, stronger personability, and—crucially—a markedly better ability to surface niche content.

These gains arise from LumiCRS’s three-pronged architecture: \textbf{(i)} the prototype-learning module for long-tail recommendation constructs context–affect–semantic prototypes for body and tail movies, stabilising sparse-item representations; \textbf{(ii)} Adaptive Comprehensive Focal Loss (ACFL) attenuates head-tier gradients while amplifying hard tail cases, reducing popularity bias; and \textbf{(iii)} the GPT-4o-based prototype-dialogue augmenter selectively injects diverse, emotionally vivid utterances around tail entities. Acting in concert, these mechanisms enable LumiCRS to produce responses that are informative, engaging, and rich in long-tail recommendations.

\section{Conclusion}
We introduced LumiCRS, a long-tail–aware conversational recommender that unifies three complementary mechanisms: (i) Adaptive Comprehensive Focal Loss (ACFL) for head suppression and tail amplification; (ii) a prototype-learning module that yields stable context–affect–semantic representations for mid- and low-frequency movies; and (iii) a GPT-4o-based prototype-dialogue augmentation pipeline that injects diverse, quality-controlled utterances.

Experiments on ReDial and INSPIRED show that LumiCRS achieves 5–10\% relative improvements in overall Recall@1/10/50, and boosts Tail-Recall@50 by 11.8\% and 12.9\%, respectively. Catalogue coverage and intra-list diversity (ILD) also increase by approximately 9\%. Dialogue generation metrics such as BLEU-2/3, ROUGE-L, and DIST-4 improve by 3–11\%, and human evaluation confirms notable gains in fluency, informativeness, persuasiveness, and long-tail relevance (TailRel +13.9\%).ese findings underscore the importance of unifying a tail-aware loss design like ACFL, prototype-level representation learning, and high-quality dialogue augmentation within a single framework. By addressing long-tail bias at the loss, representation, and data levels simultaneously, LumiCRS offers a robust and scalable solution for conversational recommendation in real-world, imbalanced scenarios.
\section{Limitations \& Discussion}
While LumiCRS delivers strong empirical gains, several limitations remain. The model depends on offline hyper-parameter tuning and GPU-intensive prototype updates, which may limit deployment in low-resource or real-time settings. Performance drops under deviations from tuned configurations suggest a lack of robustness to hyper-parameter drift. Moreover, the current evaluation is restricted to the movie domain and two English datasets, limiting conclusions about generalisation to other domains (e.g., books or e-commerce) or multilingual scenarios. This narrow scope may underrepresent the diversity and complexity of real-world recommendation environments.

In addition, while the proposed methods significantly enhance long-tail recall, the system may still struggle in scenarios involving domain shift, sparse user history, or limited dialogue context. In such cases, residual popularity bias may re-emerge, especially when the prototype module or augmentation pipeline fails to capture user intent with sufficient granularity. Moreover, the current design does not explicitly incorporate personalization signals beyond item interactions, which may limit its ability to adapt to rapidly evolving user preferences or context-dependent intents. These limitations suggest directions for improving the model’s adaptability and resilience across more diverse and dynamic conversational recommendation environments.

\printcredits

\section*{Declaration of competing interest}
The authors declare that they have no known competing financial interests or personal relationships that could have appeared to influence the work reported in this paper.
\section*{Acknowledgment}
This work was supported by the National Natural Science Foundation of China under Grant No. 61872288.
\section*{Data availability}
The data and code is publicly available at: \url{https://github.com/Jinzhi-Wang/LumiCRS-R1}
\bibliographystyle{unsrt}

\bibliography{citations}


\end{document}